\documentclass{article}
\usepackage{ijcai24}

\usepackage{times}
\usepackage{soul}
\usepackage{url}
\usepackage[hidelinks]{hyperref}
\usepackage[utf8]{inputenc}
\usepackage[small]{caption}
\usepackage{graphicx}
\usepackage{amsmath}
\usepackage{amsthm}
\usepackage{booktabs}
\usepackage{algorithm}
\usepackage{algorithmic}
\usepackage[switch]{lineno}

\usepackage{amsfonts}
\DeclareMathOperator{\E}{\mathop{\mathbb{E}}}
\DeclareMathOperator{\I}{\mathop{\mathbb{I}}}

\newcommand{\citet}[1]{\citeauthor{#1} [\citeyear{#1}]}

\title{Safety Constrained Multi-Agent Reinforcement Learning for \\  Active Voltage Control}

\author{
Yang Qu%$^1$
\and
Jinming Ma
\and
Feng Wu\thanks{Feng Wu is the corresponding author.}
\affiliations
School of Computer Science and Technology, \\
University of Science and Technology of China, Hefei, China\\
\emails
\{qu180518, jinmingm\}@mail.ustc.edu.cn,
wufeng02@ustc.edu.cn
}

\begin{document}

\maketitle

\begin{abstract}

Active voltage control presents a promising avenue for relieving power congestion and enhancing voltage quality, taking advantage of the distributed controllable generators in the power network, such as roof-top photovoltaics. While Multi-Agent Reinforcement Learning (MARL) has emerged as a compelling approach to address this challenge, existing MARL approaches tend to overlook the constrained optimization nature of this problem, failing in guaranteeing safety constraints. In this paper, we formalize the active voltage control problem as a constrained Markov game and propose a safety-constrained MARL algorithm. We expand the primal-dual optimization RL method to multi-agent settings, and augment it with a novel approach of double safety estimation to learn the policy and to update the Lagrange-multiplier. In addition, we proposed different cost functions and investigated their influences on the behavior of our constrained MARL method. We evaluate our approach in the power distribution network simulation environment with real-world scale scenarios. Experimental results demonstrate the effectiveness of the proposed method compared with the state-of-the-art MARL methods. This paper is published at \url{https://www.ijcai.org/Proceedings/2024/}.

\end{abstract}

\section{Introduction}

In recent years, significant progress has been witnessed in the development of renewable and distributed sources of electricity, such as rooftop photovoltaics (PVs) \cite{PVincrease}. While these innovations hold promise to address energy shortages and environmental concerns, they also introduce growing complexity and uncertainty into modern power systems, presenting formidable challenges. 
One of the notable challenges associated with the high penetration of distributed energy is the potential for voltage fluctuations exceeding the power grid standards \cite{PVimpact}. Mitigating these fluctuations requires harnessing control capabilities of PV inverters and other devices. As discussed in the recent work \cite{nips2021MARLavc}, achieving active voltage control across the entire network, particularly when access to global information is limited, requires intricate coordination. 

Recently, Multi-Agent Reinforcement Learning (MARL) algorithms have demonstrated exceptional performance in many domains \cite{MARLapplication,WWprima20}, and MARL algorithms are gradually finding applications in real-world scenarios. In comparison to traditional voltage regulation methods like Optimal Power Flow (OPF) \cite{OptimalPower,7042735} and droop control \cite{droop}, MARL algorithms offer several advantages:
1) They follow the Centralized Training with Decentralized Execution (CTDE) schema, and correspond to the trait that usually only limited local information can be accessed in active voltage control. 
2) They are data-driven and are naturally model-free, while traditional methods need exact system models. 
3) Traditional methods have high computational complexity in solving power flow equations, having difficulty in real-time response. Meanwhile, MARL shows adaptability to respond to environmental changes quickly. 

To date, there have been some previous efforts to apply MARL on voltage control problems and achieved promising performances \cite{powerRLsurvey}. 
Previous research has explored the potential of MARL by taking various approaches, such as integrating traditional control techniques \cite{9076841,9328796}, enhancing state information representation and region segmentation \cite{cao2020distributed}, refining MARL reward designs \cite{nips2021MARLavc,9076841} and so on, achieving remarkable advances in the field. 
However, the unique property of constrained optimization of this problem is less concerned in the previous work. 
Specifically, in the active voltage control problem, the constraint is the voltage threshold and the objective is the total power loss.
Moreover, unlike most safe RL benchmarks \cite{safegym}, the fulfillment of the constraint is the primary concern. 
In the power distribution network, large voltage fluctuations will affect the stability of the power system, reduce power quality for users, and even cause irreversible damage to the system equipment, resulting in system failure and collapse. 
It is relatively acceptable to ensure the safety of the system at the cost of higher power loss. 
Given the dynamic of the system, it is generally more difficult and complicated to meet the constraint than to improve the objective in this problem, for the constraints must be fulfilled given a sequence of uncontrollable events. 
Many real-world problems exhibit the aforementioned two features, for example autonomous driving, wireless security, traffic control \cite{MWitits23} and so on. 
To the best of our knowledge, the application of constrained RL on these problems is less studied in the literature \cite{safeRLreview}. 

Against this background, we propose a novel constrained MARL approach named Multi-Agent RL with Double Estimation of Lagrangian Constraint (MA-DELC). 
Specifically, we formulate the active voltage control problem as a constrained Markov game. 
Firstly, we incorporate a safety critic and a cost estimator alongside the conventional reward critic. 
The safety critic is used for guiding the policy to fulfill the constraint in the long term, and the cost estimator is used for updating the Lagrange-multiplier. 
These additional components allows our system to effectively balance the optimization objectives and the imposed constraints. 

The results of our experiments highlight the efficacy of this approach, demonstrating that the incorporation of structural information significantly enhances agent performance. 
Additionally, we explore the conversion of voltage constraints into various cost functions, meticulously studying the impact of different cost function designs through comprehensive experimental analysis. 
This exploration is of great practical significance, particularly when applying constrained MARL techniques to real-world applications, 
where the choice of cost function can have a profound influence on algorithm behavior and performance. 
Our experiments on the MAPDN environment \cite{nips2021MARLavc}, commonly used in the literature, show that these components enhance performance and scale well compared with the state-of-the-art RL methods. 

\section{Related Work} \label{relatedwork}

Here, we briefly review recent MARL and constrained RL approaches for active voltage control. 

\subsection{MARL for Active Voltage Control}

Recently, MARL algorithms have gained significant attention in active voltage control. 
\citet{9076841} and \citet{9328796} combined traditional voltage control methods and MARL.
\citet{9076841} used MADDPG \cite{MADDPG} with a manually designed voltage inner loop, 
and agents set reference voltage instead of reactive power as their control actions. 
\citet{9328796} proposed a two-stage volt-var control method, in which the traditional optimal ﬂow method \cite{OptimalPower} is used to dispatch on-load tap 
changer and capacitor banks in the first stage, 
and in the second stage the MADDPG algorithm is used to regulate the reactive power of PVs. 
\citet{9353702} adopted a discrete action space of the set of all tap positions of the voltage regulating device, 
and proposed a consensus-based maximum entropy MARL framework composed of max-entropy MARL algorithm and a consensus strategy. 
\citet{cao2020distributed} used spectral clustering to divide the power distribution network into several sub-networks 
according to the voltage and reactive power sensitivity, and then applied MATD3 \cite{MATD3} to solve the voltage control problem, each sub-network corresponds an agent. 
\citet{cao2020distributed} also used reactive power as control actions. 
\citet{9399637} used both reactive power and the curtailment of active power as control actions, and applied MASAC \cite{MASAC} to solve decentralized voltage control problems with high penetration of PVs. 
\citet{9113746} and \citet{10.1145/3534678.3539480} combines MADDPG \cite{MADDPG} critic with transformer to enhance inter-agent coordination. 

In this paper, we adopt continuous control action space of reactive power of PV inverters, which is more flexible and reliable than discrete control action. It is worth noting that our approach does not need topological information, and maintains distributed during execution. 

\subsection{Constrained RL for Active Voltage Control}

Now, we discuss the methods that applied constrained RL algorithms to solve the active voltage control problem. 

\citet{9867476} designed a stability-constrained RL framework which utilizes a manually designed Lyapunov function, 
and proved stability guarantees. 
\citet{8909741} directly applied Constrained Policy Optimization (CPO) \cite{CPO} to solve the Volt-VAR problem. 
\citet{10029892} utilized a safety layer to keep the action within safety limits 
by calculating the power ﬂow equations based on the physical model of the power grid. 
The above methods are all based on single-agent settings, and are trained and executed in a centralized style. 

\citet{IEEE-CMASAC} formulated the Volt-VAR Control problem as a constrained Markov game 
with the reactive power of inverters or static var compensators (SVCs) as control actions, 
and proposed a Multi-Agent Constrained Soft Actor-Critic algorithm (MACSAC) that combines MASAC \cite{MASAC} and Lagrangian constraint. 
\citet{AAMAS-safetylayer} used safety layer method based on multi-agent settings, and they used neural networks to approximate cost functions, getting rid of the need of precise physical models. 

Although \citet{IEEE-CMASAC} also used the Lagrangian constraint MARL method, our approach differs in the double safety estimation framework and the adaptive Lagrange multiplier. Actually, MACSAC is similar to MADELC w/o cost estimator in our ablation study in Section 5.3.
\citet{AAMAS-safetylayer} focuses on the safety issue and ignores power loss reduction. Moreover, it considered only one-step cost, while our approach considers both one-step cost and the cumulative cost return, enabling agents to minimize costs more steadily. 

\section{Background} \label{formulation}

\subsection{Active Voltage Control on Power Networks}

\begin{figure}[t]
        \centering
        \includegraphics[width=\linewidth]{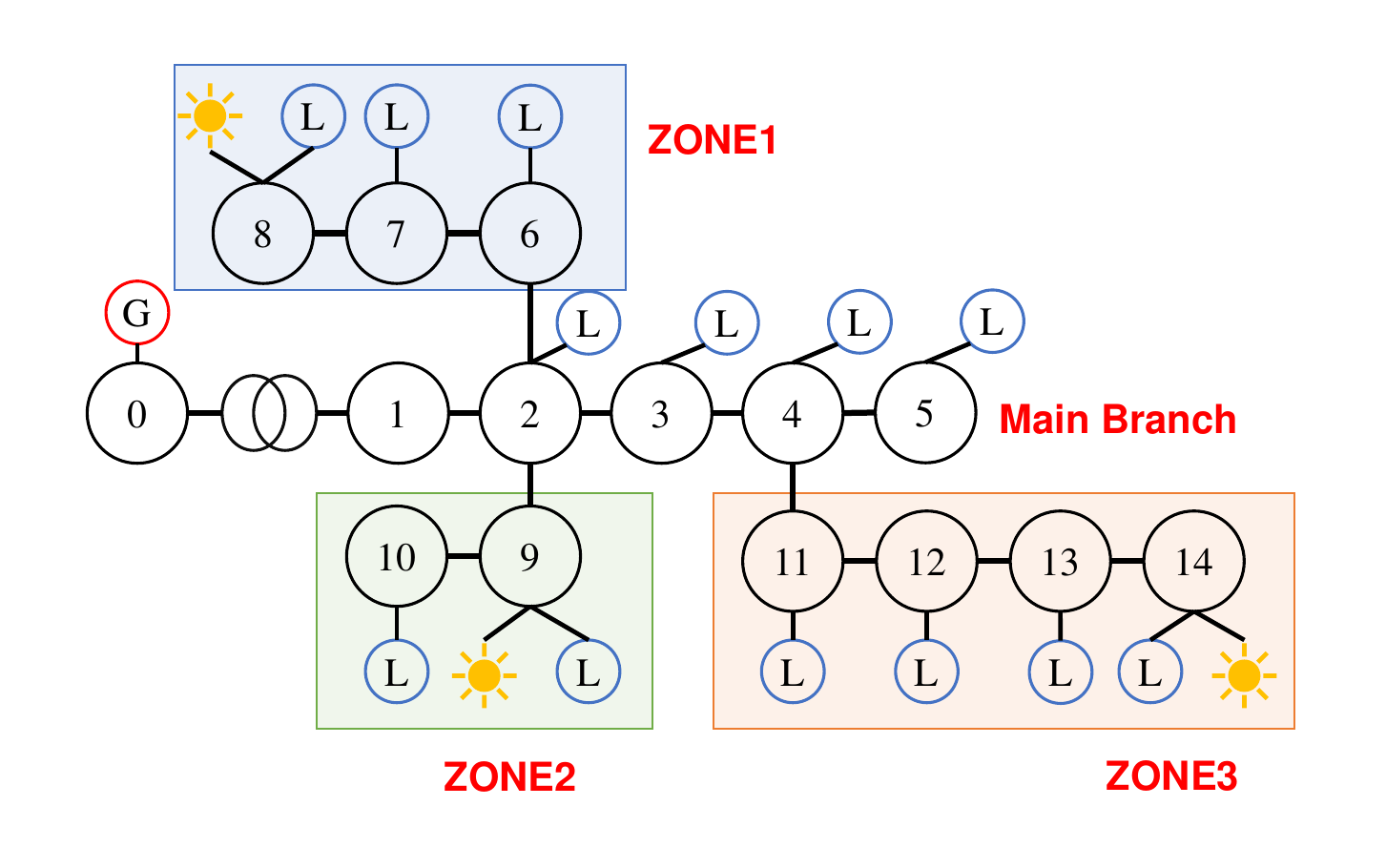}
        \caption{An example of power distribution networks. The black circles with numbers represent buses; 
        G represents an external generator; Ls represent loads; and the sun emoji represents the location where a PV is installed.
        We control the bus voltage by adjusting the reactive power generation of PV inverters in buses 8, 9, and 14. }
        \label{fig:powergrid}
\end{figure}

% model powergrids as tree graphs
In this paper, we consider a power distribution network installed with roof-top photovoltaics (PVs).
As shown in Figure~\ref{fig:powergrid}, we model the power distribution network as a tree graph $\mathcal{G}=(V, E)$, where $V=\{0,1,\dots,|V|\}$ represents the set of nodes (buses), and $E=\{1, 2,\dots,|V|\}$ represents the set of edges (branches).
Nodes in the distribution network are divided into several zones based on their shortest path from the terminal to the main branch \cite{OptimalPower}. 
We denote regions as Zone$_z, z=1,\dots,|Z|$. 
Node 0 is connected to the main grid, and has a stable voltage magnitude, denoted as $v_0$. 
Each node may be connected to load and PV. 
We denote the set of nodes installed with PVs in Zone$_z$ as PV$_z$, and the set of nodes connected with load in Zone$_z$ as Load$_z$. 

For each bus $j \in V$ , let $v_j$ and $\theta_j$ be the magnitude and phase angle of the voltage, 
and $s_j = p_j + \mathbf{i}q_j$ denotes the complex power injection. 
Note that there are complex and non-linear relationships between these physical quantities that satisfy the power system dynamics rules \cite{PowergridDynamics,nips2021MARLavc}. Hence, it is hard to solve it in close form.

% Active Voltage Control: a constrained optimal problem 
Let $v_0 = 1.0$ per unit ($p.u.$). For other buses, 5\% voltage deviation is usually allowed for safe and optimal operation.
The objective of active voltage control is to keep the nodal voltages in a normal range (e.g., 0.95 $p.u.$ to 1.05 $p.u.$) 
and concurrently minimize total active power loss $P_{loss} = \Sigma _{(i,j)\in E}R_{ij}I_{ij}^2$, or known as the line loss. 
$R_{ij}$ is the resistance of branch $(i,j)$ and $I_{ij}$ is the current magnitude from bus $i$ to $j$ \cite{9805763}. The optimization problem is as follow:
\begin{equation}\label{eq:avc}
\begin{array}{ll}
\min & P_{loss} = \sum_{(i,j)\in E}R_{ij}I_{ij}^2\\[5pt]
\mbox{s.t.} & 0.95 p.u. \le v_i \le 1.05 p.u., \forall i \in V\backslash \{0\}; \\[2pt]
& \mbox{satisfy power flow dynamics}. 
\end{array}
\end{equation}

When loads are heavy at nighttime, $v_i$ may drop below 0.95$p.u.$. Meanwhile, large power generation of PVs in the middle of the day may cause $v_i$ rise to exceed 1.05$p.u.$.
These problems can be handled by applying appropriate control on the reactive power injection of the PV inverter.

\subsection{Constrained Markov Game for Active Voltage Control}

We consider the control process of the PV inverters as constrained Markov game \cite{CMG}: $\mathcal{M} = \langle \mathcal{N}, \mathcal{S}, \{\mathcal{O}_i\}_{_i \in \mathcal{N}}, \{\mathcal{A}_i\}_{_i \in \mathcal{N}}, \mathcal{T}, r, \Omega, \mathcal{C}, c, \gamma \rangle$, 
where $\mathcal{N}=\{1,2,\dots,n\}$ is a set of $n$ agents; 
$\mathcal{S}$ denotes the state space; 
$\mathcal{O}=\times_{i\in\mathcal{N}}\mathcal{O}_i$ denote the joint observation set, where $\mathcal{O}_i$ denotes the observation set of agent $i$; 
$\mathcal{A}=\times _{i\in\mathcal{N}}\mathcal{A}_i$ denote the joint action set, where $\mathcal{A}_i$ denotes the action set of agent $i$; 
$\mathcal{T}:\mathcal{S}\times\mathcal{A}\times\mathcal{S}\rightarrow [0,1]$ is probabilistic state transition function; 
$r$ is the reward function;
$\Omega: \mathcal{S}\times\mathcal{A}\times\mathcal{O}\rightarrow [0,1]$ denotes the perturbation of the observers for agents' joint observations over the states after decisions; 
$\mathcal{C}:\mathcal{S}\times\mathcal{A}\rightarrow\mathcal{R}$ is 
the cost function; 
$c$ denote the cost limit. 
MARL algorithms for constrained Markov game aim to search a policy $\pi$ and solve this constrained optimization problem: 
\begin{equation}\label{eq:cmdp}
\begin{array}{ll}
        \max_{\pi} & \E_{(s_t, a_t)\sim \pi}\left[\sum_t\gamma^tr(s_t,a_t)\right], \\[5pt]
        \mbox{s.t.} & \E_{(s_t, a_t)\sim \pi}\left[\sum_t\gamma^t\mathcal{C}_i(s_t,a_t)\right] < c_i, i=1,2,\dots,m.
\end{array}
\end{equation}

In the context of the active voltage control problem within the Constrained Markov Game framework, the essential elements are defined as follows:
\begin{itemize}
  \item \textbf{Agent}: Each agent is responsible for controlling an individual PV inverter in the system. 
  \item \textbf{Observation}: The observation of an agent is composed of relative physical quantities of the buses in the region to which the agent belongs. 
  Specifically, $\mathcal{O}_i=\mathcal{L}_r\times\mathcal{P}_r\times\mathcal{V}_r$, 
  where $\mathcal{L}_r=\{p^L_j, q^L_j>0|\forall V_j \in \text{Load}_r \}$: $p^L_j, q^L_j$ are 
  active and reactive power of the load connected to node $j$ respectively; 
  $\mathcal{P}_r=\{p^{PV}_j, q^{PV}_j>0 | \forall V_j \in \text{PV}_r \}$: $p^{PV}_j, q^{PV}_j$   are active and reactive powers generated by PV inverters in node $j$; 
  $\mathcal{V}_r=\{v_j>0, \theta_j \in [-\pi,\pi] | \forall V_j \in \text{Zone}_z\}$, $v_j, \theta_j$ is voltage magnitude and phase of node $j$. 
  Agents in the same region share the same observation. 
  \item \textbf{Action}: Each agent $i\in\mathcal{N}$ has a continuous action set $ \mathcal{A}_i= \{a_i:-1.0\le a_i\le 1.0\}$ 
  that denotes the ratio of maximum reactive power it can generate. 
  \item \textbf{State and State Transition}: The global state is composed of bus voltage, the active and reactive power of loads and PVs of all nodes. 
  Power of loads and active power generation of PVs of the next state are obtained from the dataset, and other parts are calculated by solving power dynamic equations with data from the dataset, current state and action. 
  \item \textbf{Reward}: Agents receive reward at each time step: $r = - l_q(q^{PV})=-\frac 1 {|I|} ||q^{PV}||$ , which is the mean reactive power generation loss of PVs, and is an approximation of power loss caused by PVs.
  \item \textbf{Constraint}: According to Eq.\ref{eq:avc}, the voltage safety constraint corresponds to the cost function $\mathcal{C}(s, a)=\I((v_i<0.95||v_i>1.05) | \exists i \in V\backslash \{0\})$. 
  We also propose other cost functions in order to provide more information and guide the agents, which are described in section \ref{Cost-functions}. During training, the algorithm utilizes normalized rewards and costs, with cost values normalized to the range (-1,1). We expect that all bus voltages remain within the safe range, and the cost limit $c$ is set to -0.5 under the normalized standard.
\end{itemize}

\section{METHOD} \label{method}

We propose the Multi-Agent reinforcement learning with Double Estimation of Lagrangian Constraint (MA-DELC) algorithm 
and study the impact of different cost function designs. 
These components are introduced below in details. 

\subsection{MARL with Double Estimation of Lagrangian Constraint}

% Inspired by SAC-Lagrangian\cite{SACLagrange}, we extend this approach to multi-agent settings.
Constrained optimization problems in Eq.~\ref{eq:cmdp} can be solved by the Lagrange-multiplier method \cite{LagrangeMultiplier}. 
Specifically, it introduces a Lagrangian function $\mathcal{L}(\pi,\alpha)$ as defined:
\begin{equation}\label{eq:cmdp-lagrange}
\begin{split}
        \mathcal{L}(\pi,\alpha) & \doteq f(\pi)-\alpha (g(\pi)-c), \mbox{where} \\
        f(\pi) & = \E_{(s_t, a_t)\sim \pi} \left[ \sum_t\gamma^tr(s_t,a_t) \right], \mbox{and} \\
        g(\pi) & = \E_{(s_t, a_t)\sim \pi} \left[ \sum_t\gamma^t\mathcal{C}(s_t,a_t) \right].
\end{split}
\end{equation}
where $\alpha>0$ is the Lagrange-multiplier or the dual variable; $f(\pi)$ is the discounted cumulative return of reward; 
$g(\pi)$ is the discounted cumulative return of cost; and $c$ is the cost limit. 
Here, $\mathcal{L}(\pi,\alpha)$ can be viewed as the optimization objective, 
and coefficient $\alpha$ determines the degree of emphasis on the constraint $g(\pi)$.

In MA-DELC, two separate critics (Q functions) are trained, where one is a reward-critic $Q^r_\phi$ for estimating the cumulative reward return $f(\pi)$, 
and the other one is a cost-critic $Q^c_\phi$ for estimating $g(\pi)$, $\phi$ denote the parameters of the critic networks. 
Follow the formulation of SAC-Lagrange \cite{SACLagrange} and MADDPG \cite{MADDPG}, we use TD-error loss to train critics: 
\begin{equation}\label{eq:critic_loss}\small
\begin{split}
J_{Q}(\phi) &= \E_{(s_t, a_t)\sim\mathcal{D}} \left[(Q^r_{\phi}(s_t, a_t)-y^r)^2+(Q^c_{\phi}(s_t, a_t)-y^c)^2 \right], \\
y^r &= r_t+\gamma \bar{Q}^r_{\phi '}(s_{t+1},a_{t+1})|_{(r_t,s_{t+1})\sim\mathcal{D},a_{t+1}\sim\pi_{\theta '}(s_{t+1})}, \\
y^c &= c_t+\gamma \bar{Q}^c_{\phi '}(s_{t+1},a_{t+1})|_{(c_t,s_{t+1})\sim\mathcal{D},a_{t+1}\sim\pi_{\theta '}(s_{t+1})}.
\end{split}
\end{equation} 
$\bar{Q}^r_{\phi '}, \bar{Q}^c_{\phi '}$ denotes the target critic network with the parameter $\phi'$ that is soft copied from the critic network every $\tau$ steps. 

In existing primal-dual constrained RL methods \cite{SACLagrange,PDO}, the cost-critic $Q^c$ is directly used for updating the dual variable, 
which is inaccurate due to the uncertainty of cumulative cost and the bias of the Q value approximation. 
This may lead to the algorithm overestimating the constraint satisfaction. %, as we will show in the results of our ablation experiment. 
To avoid these errors, we train a one-step cost estimator $\hat{C}$ with the parameter $\psi$ for adjusting the dual variable $\alpha$ with the mean square error. 
\begin{eqnarray}\label{eq:hat_C-loss}
        J_{\hat{C}}(\psi) = ||\hat{C}(s_t, a_t)-c_t||^2_2|(s_t, c_t)\sim\mathcal{D},a_t\sim\pi_\theta(s_t)
\end{eqnarray}

On the purpose of maximizing the Lagrangian function $\mathcal{L}(\pi,\alpha)$, we can get the actor loss:
\begin{eqnarray}\label{eq:actor-loss}
        J_{\pi}(\theta) = \E_{s_t\sim\mathcal{D},a_t\sim\pi_{\theta}} \left[
        -Q^r_{\phi}(s_t, a_t)+\alpha Q^c_{\phi}(s_t, a_t) \right]
\end{eqnarray}
\setlength{\arraycolsep}{5pt}
where $\mathcal{D}$ is the replay buffer and $\theta$ is the parameters of the actor network. 

To ensure that the constraint $g(\pi)<c$ is satisfied, 
we learn the adaptive safety weight $\alpha$ (Lagrangian multiplier) by minimizing the loss $J(\alpha)$:
\begin{equation}\label{eq:alpha-update}
        J(\alpha)=\E_{s_t\sim\mathcal{D},a_t\sim\pi_{\theta}}\left[ \alpha(c-\hat{C}_{\psi}(s_t, a_t)) \right]
\end{equation}
In this case, if $\hat{C}_{\psi}(s_t, a_t)>c$, the constraint is not satisfied, so $\alpha$ 
will be increased to emphasize safety more; otherwise $\alpha$ will be decreased.

\begin{algorithm}[t]
  \caption{Training process of MA-DELC}
  \label{alg:MA-DELC}
  \begin{algorithmic}[1]
    \STATE Initialize replay buffer $\mathcal{D}$, actor and critic parameters.
    \FOR {each episode }
        \FOR {each time step $t$ }
            \FOR {agent $i\in\mathcal{N}$}
                \STATE Select action $a_i=\mu_{\theta_i}(o_i)+\xi$, $\xi\sim\mathcal{N}(0,\Sigma_{std})$.
            \ENDFOR
            \STATE Execute joint action $a=(a_1,\dots,a_n)$ at state $s$.
            \STATE Get reward $r$, cost $c$ by going to next state $s'$.
            \STATE $\mathcal{D}\leftarrow\mathcal{D}\cup (s,a,r,c,s')$, $s\leftarrow s'$.
            \STATE Sample batch $\mathcal{B}$ from $\mathcal{D}$.
            \STATE Update $Q^r_\phi, Q^c_\phi$ by minimizing the loss in Eq.~\ref{eq:critic_loss}.
            \STATE Update $\hat{C}_\psi$ by minimizing the loss in Eq.~\ref{eq:hat_C-loss}.
            \STATE Update $\pi_\theta$ by minimizing the loss in Eq.~\ref{eq:actor-loss}.
            \STATE Update  $\alpha$ by minimizing the loss in Eq.~\ref{eq:alpha-update}.
            \STATE $\theta '\leftarrow\tau\theta ' + (1-\tau)\theta ,\phi '\leftarrow\tau\phi ' + (1-\tau)\phi$ .
        \ENDFOR
    \ENDFOR
     \end{algorithmic}
\end{algorithm}

\begin{figure}[t]
        \centering
        \includegraphics[width=.5\linewidth]{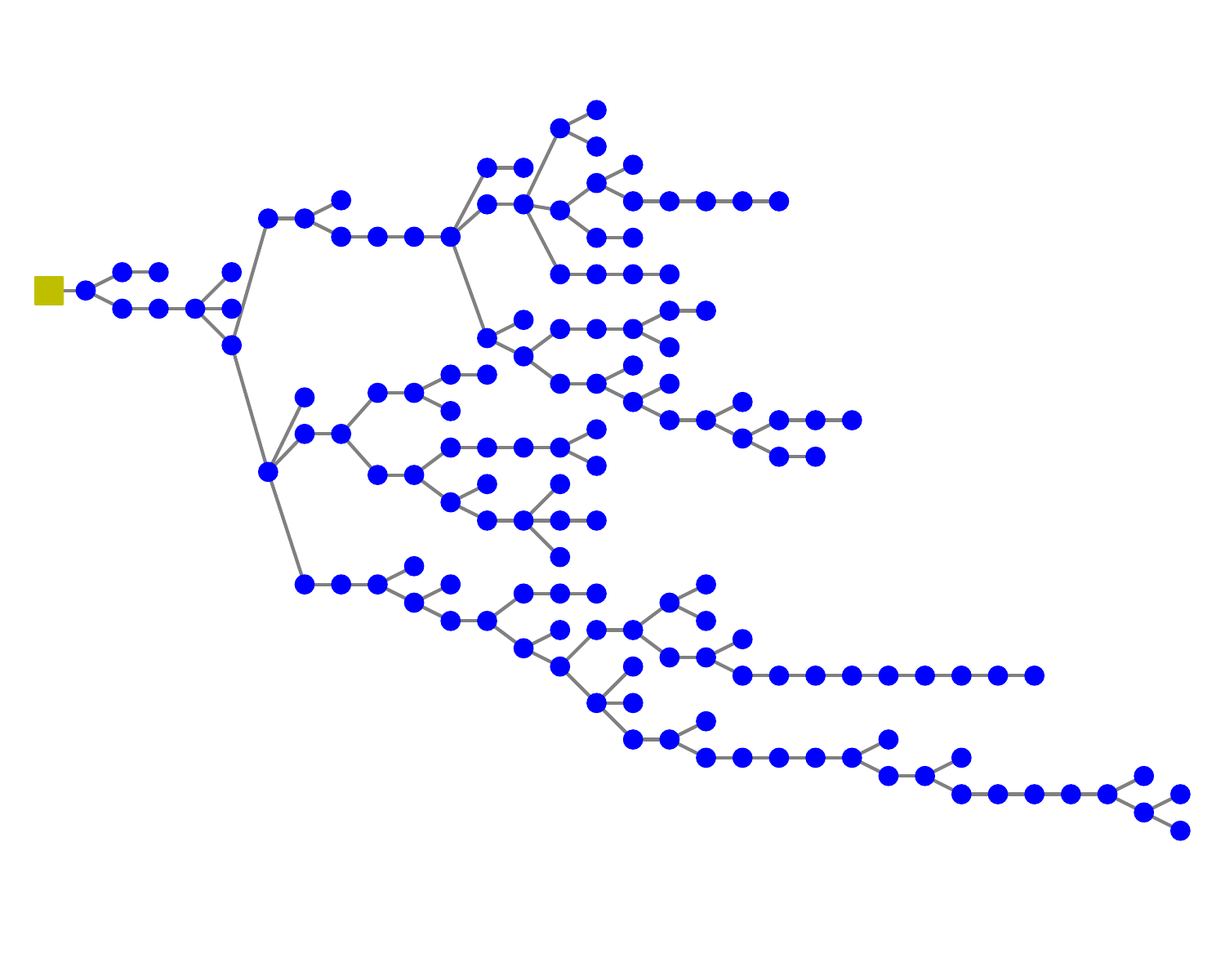}\hfill
        \includegraphics[width=.5\linewidth]{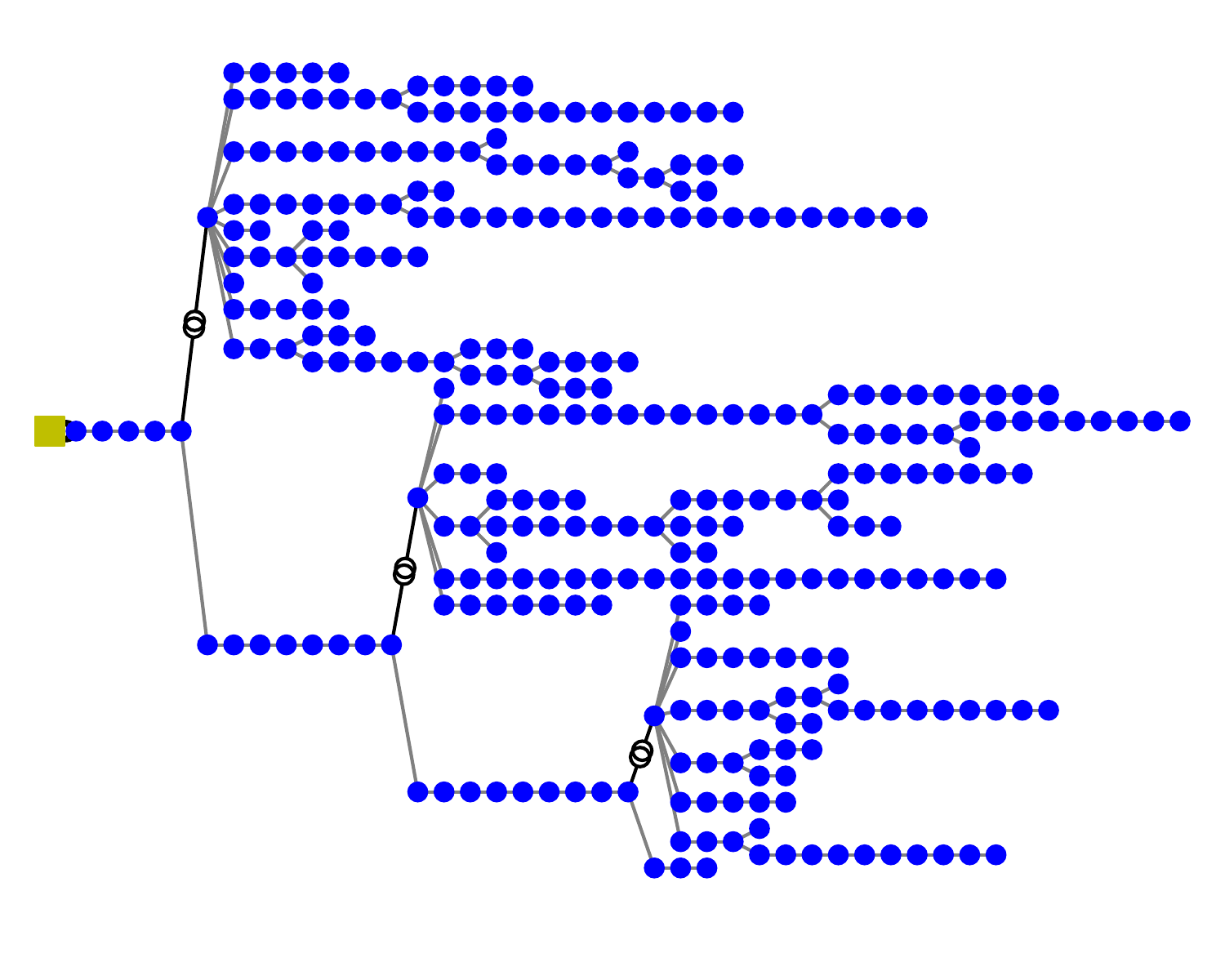}
        \caption{The topology of the distribution network of the 141-bus and 322-bus scenario in the MAPDN environment.}
        \label{fig:network}
\end{figure}

\begin{figure*}[ht]	
 \begin{minipage}{0.3\linewidth}
 	\vspace{3pt}
 	\centerline{\includegraphics[width=\textwidth]{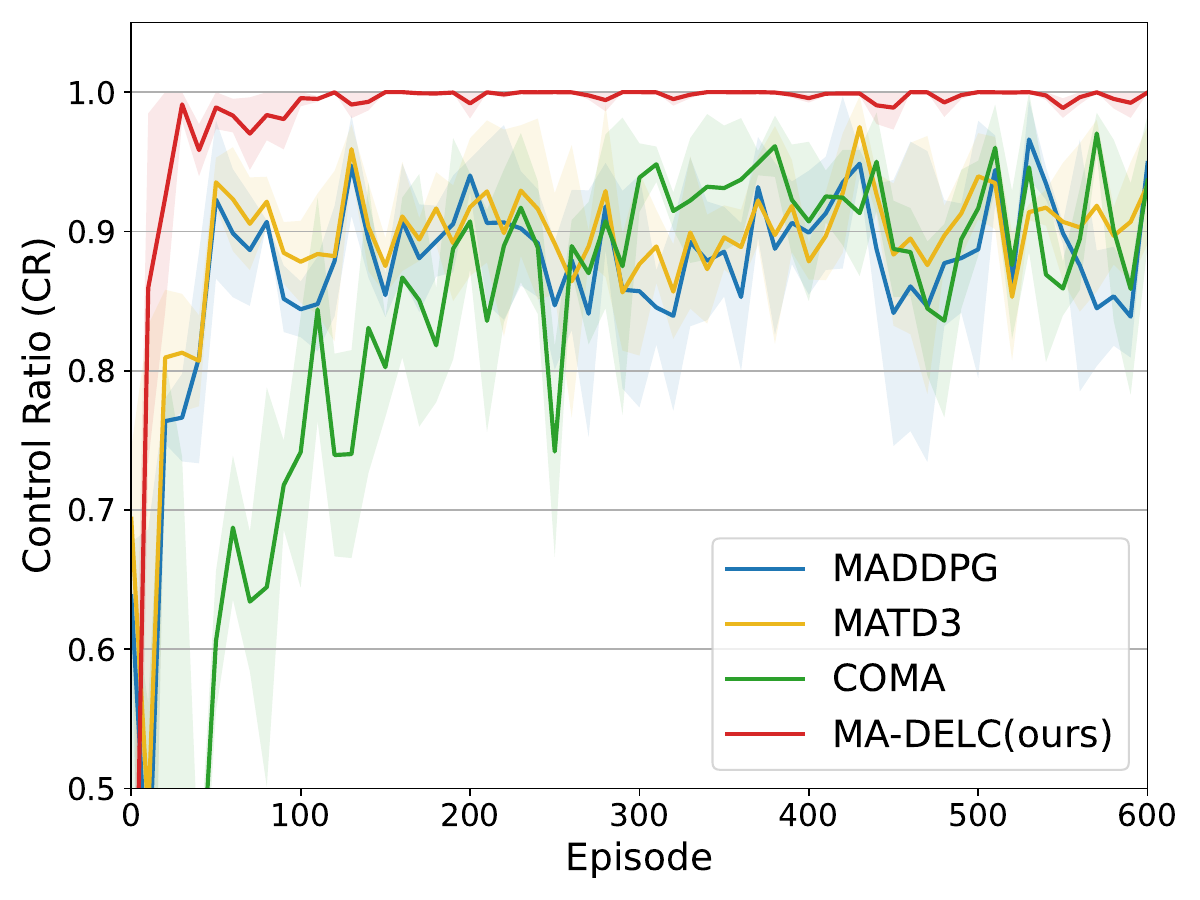}}
 	\vspace{3pt}
 	\centerline{\includegraphics[width=\textwidth]{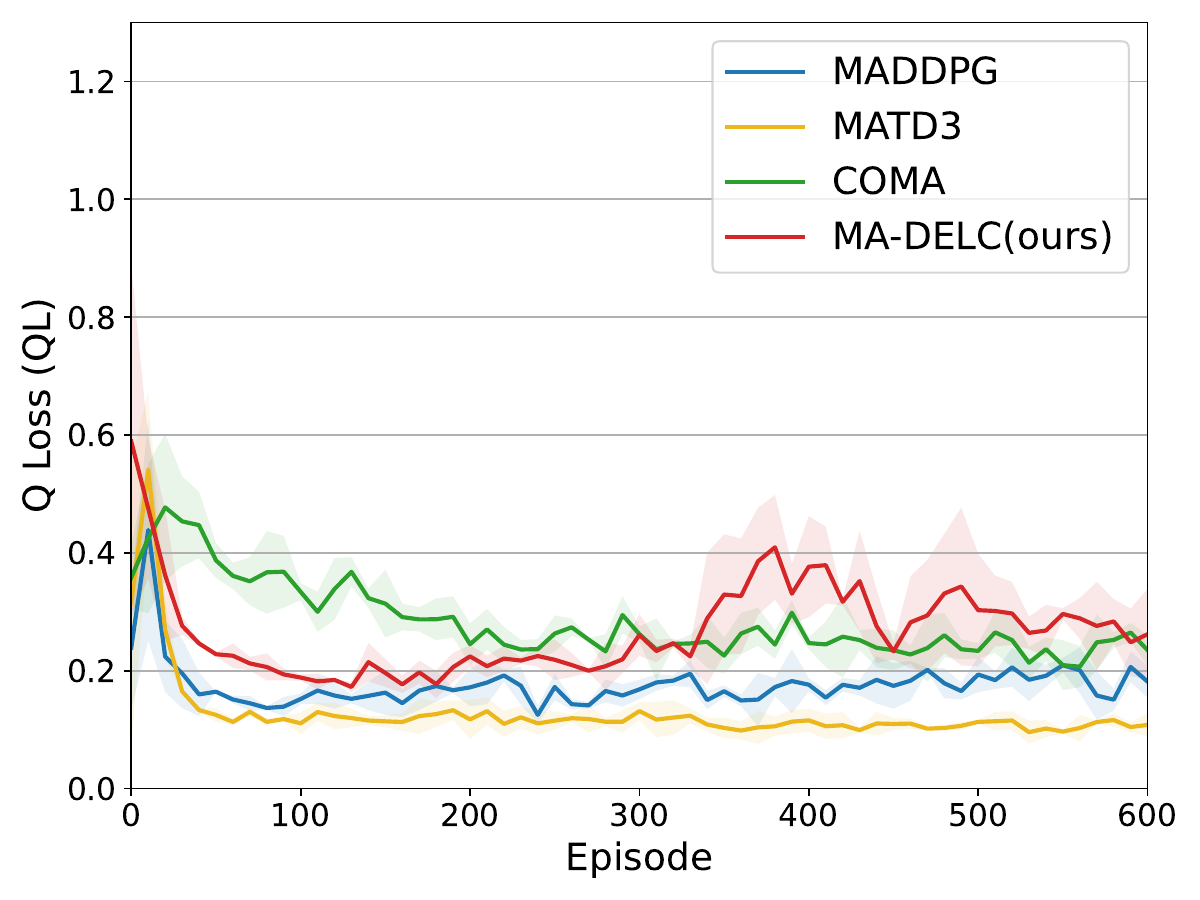}}
 	\vspace{3pt}
 	\centerline{(a) 33-bus scenario}
 \end{minipage}\hfill
 \begin{minipage}{0.3\linewidth}
	\vspace{3pt}
 	\centerline{\includegraphics[width=\textwidth]{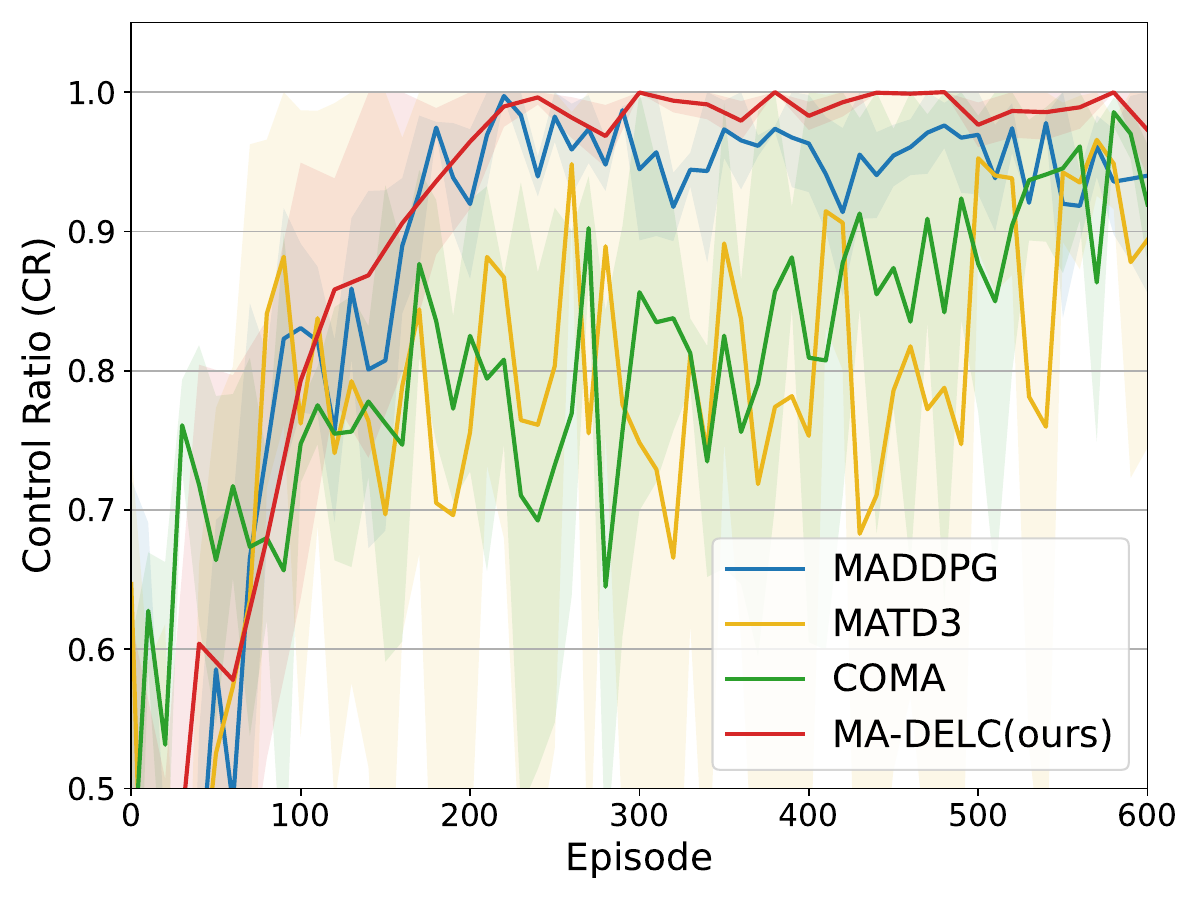}}
 	\vspace{3pt}
 	\centerline{\includegraphics[width=\textwidth]{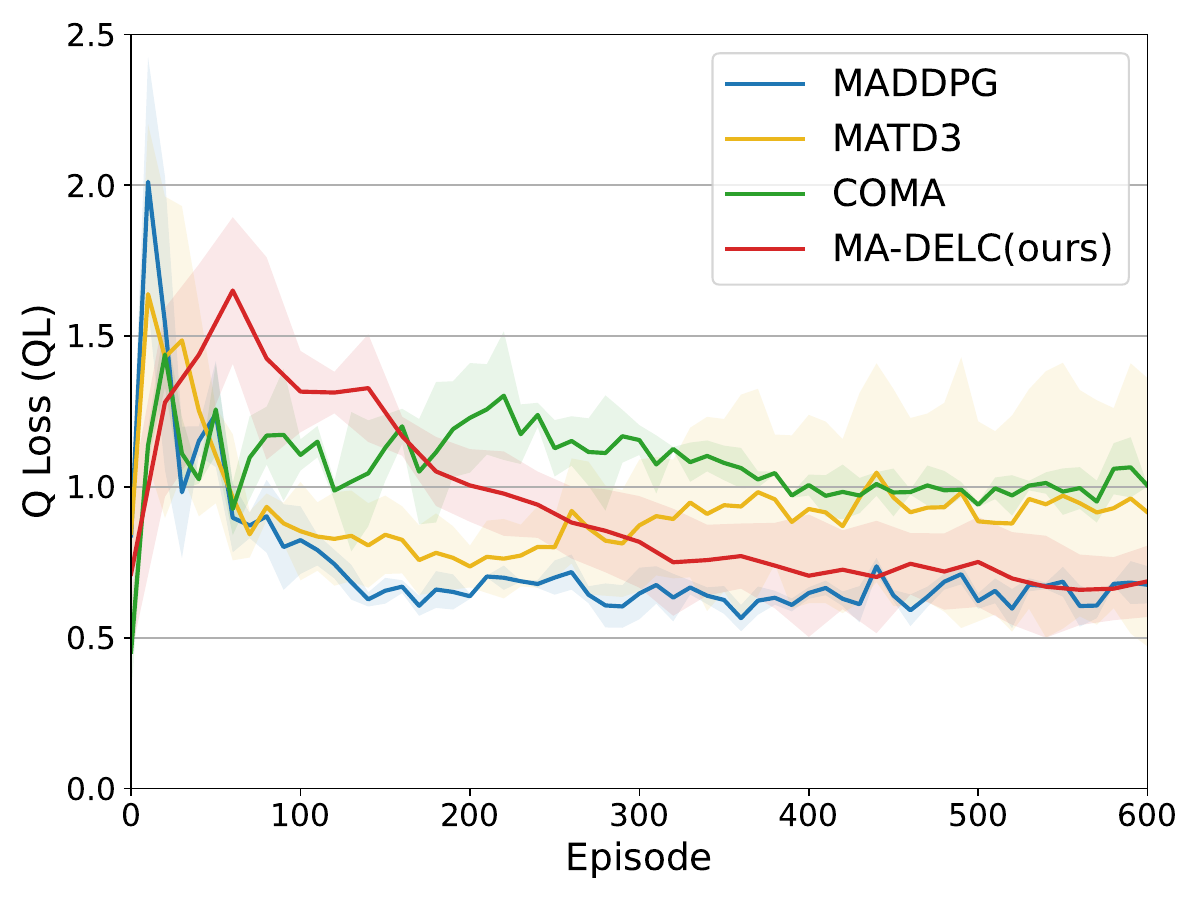}}
 	\vspace{3pt}
	\centerline{(b) 141-bus scenario}
\end{minipage}\hfill
\begin{minipage}{0.3\linewidth}
	\vspace{3pt}
 	\centerline{\includegraphics[width=\textwidth]{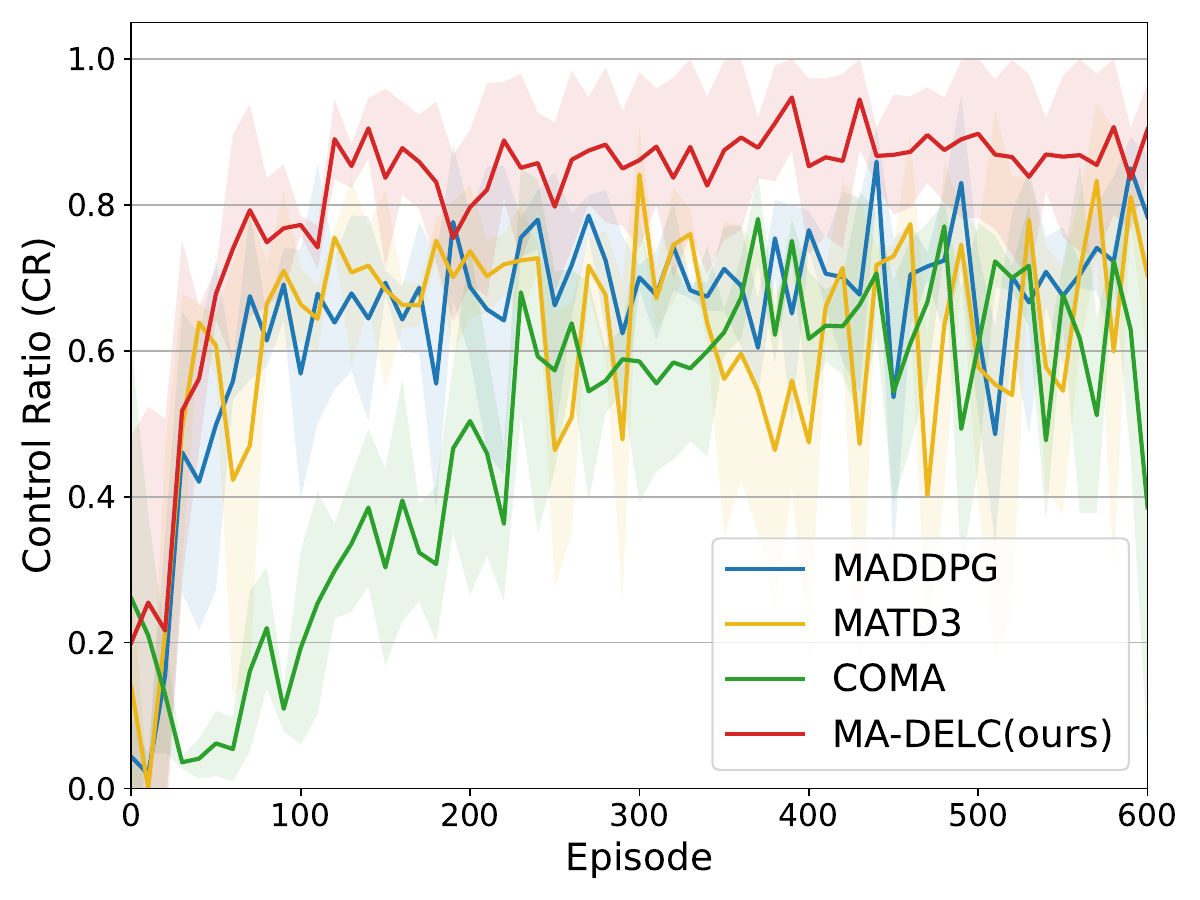}}
 	\vspace{3pt}
 	\centerline{\includegraphics[width=\textwidth]{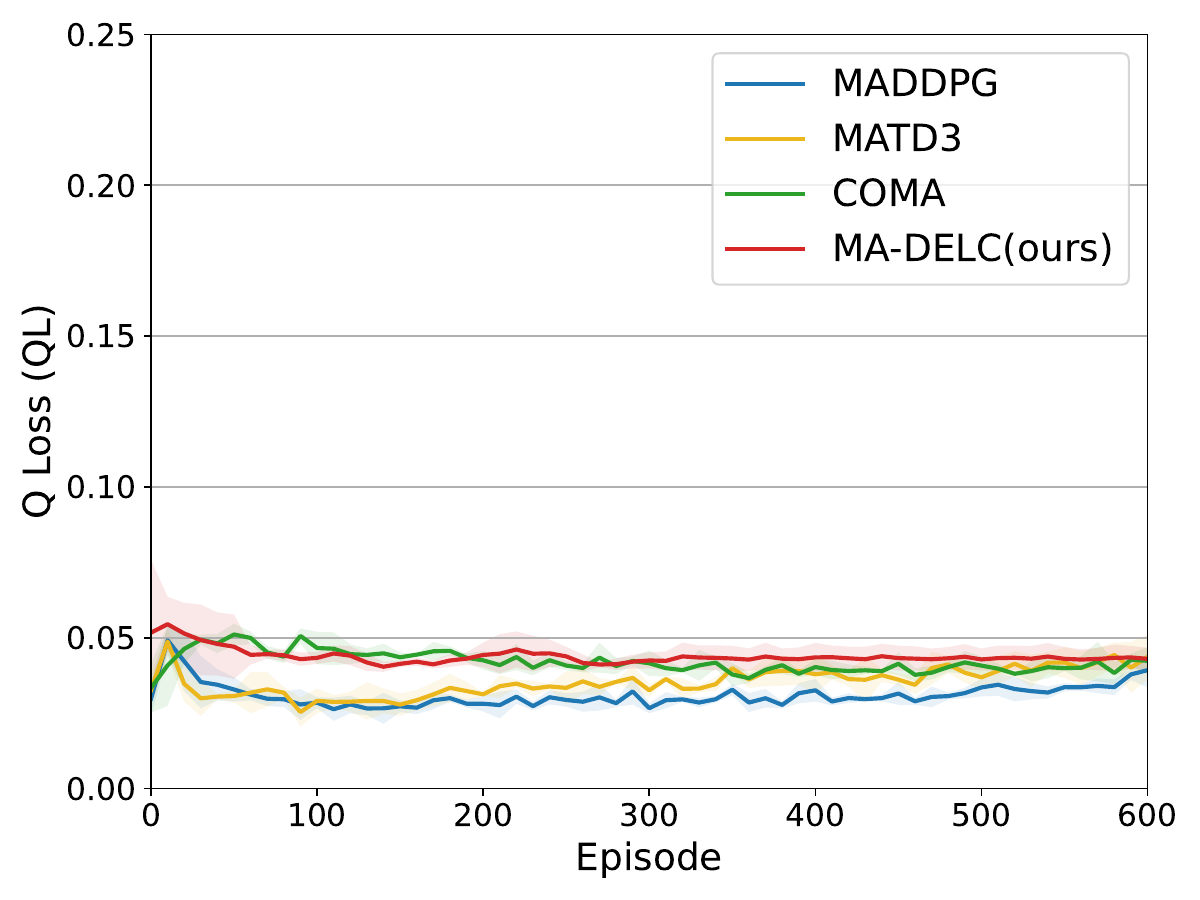}}
 	\vspace{3pt}
	\centerline{(c) 322-bus scenario}
 \end{minipage}
 \caption{Control Ratio (CR) and Q Loss (QL) results of the different algorithms (CR: higher is better, QL: lower is better). %Results of MA-DELC in the 322-bus scenario used step cost function.
 }
 \label{fig:result}
\end{figure*}

\subsection{Cost Functions} \label{Cost-functions}

In Section \ref{formulation}, we formalize the cost function according to the form of the constraint and the Lagrange-multiplier method. 
However, this boolean cost function is not informative enough for agents to learn from in large-scale scenarios. 
To provide more information about the constraint and the degree of violation, we propose several cost functions below:
% \mathcal{C}(s, a)=\mathbf{I}(\exists i (v_i<0.95||v_i>1.05))
\begin{enumerate}
\item \textbf{Boolean Cost}: This is the basic cost function that implies whether the constraint is satisfied. 
If all buses' voltages are in the safe range at the time step, the cost is $0$; otherwise, the cost is $1$. 
Denote 
\begin{equation}
    C_{percent} = 1-\frac{\sum_i \I(v_i<0.95 || v_i>1.05)}{|V|}
\end{equation}
as the percent of buses whose voltages are within the safety limit. 
Then this boolean cost function is as: 
\begin{equation}
        \mathcal{C}(s, a)=\I(C_{percent} = 1.0).
\end{equation}
\item \textbf{Step Cost}: We construct the piece-wise cost function to provide more information about the whole picture of the power grid. 
This cost function differs from the boolean cost function in that it returns a lower cost when more than 90\% of all buses are under control. 
We expect the step cost function to help the agents to distinguish between safe and unsafe actions when the constraint is hard to be fulfilled completely 
and the agents cannot get information from the boolean cost function. 
\begin{equation}
        \mathcal{C}(s, a) = \left\{
                \begin{aligned}
                0, & \quad C_{percent} = 1.0\\
                0.5, & \quad C_{percent} \in [0.9,1.0)\\
                1, & \quad C_{percent} \in [0, 0.9)\\
                \end{aligned}
                \right
                .
\end{equation}
\item \textbf{V-Loss Cost}: The v-loss cost function measures the mean voltage deviation from $1.0p.u.$ of all buses. 
This cost function is continuous and is the most informative one among other cost functions. 
Minimizing v loss means keep nodal voltage as stable as possible, and correlates with safety. 
However, it does not give explicit information about whether the constraint is violated. 
\begin{equation}
        \mathcal{C}(s, a)=\frac{1}{|V|}\sum_{i\in V}l_v(v_i)=\frac{1}{|V|}\sum_{i\in V}(|v_i-1.0|_1)
\end{equation}

\end{enumerate}

\subsection{Implementation Details}

The main procedures of our MA-DELC method are outlined in Algorithm~\ref{alg:MA-DELC}. 
In lines 4-9, agents interact with the environment to obtain transition data. 
We add noise $\xi$ on the action to be executed to encourage exploration. 
The variance $\Sigma_{std}$ of the noise $\xi$ is a hyper-parameter. 
Actors of different agents use shared parameters, and the input of actors is local observation $o_i$ appended with agent id. 
Next, we update $\alpha$ and the parameters of actor, critic, cost estimator and target critic networks in lines 10-15. 

% \begin{figure*}[t]
% 	\centering
% 	\subfigure[1]
% 	{\includegraphics[width=.8\columnwidth]{fig/CR-33bus}}
% 	\subfigure[2]
% 	{\includegraphics[width=.8\columnwidth]{fig/CR-141bus}}
%     \subfigure[3]
% 	{\includegraphics[width=.8\columnwidth]{fig/CR-322bus}}
% 	\caption{}
% 	\label{ablation}
% \end{figure*}

\section{Experiments} \label{experiments}

\begin{figure*}[t]
\centering
 \begin{minipage}{0.3\linewidth}
 	\vspace{3pt}
 	\centerline{\includegraphics[width=\textwidth]{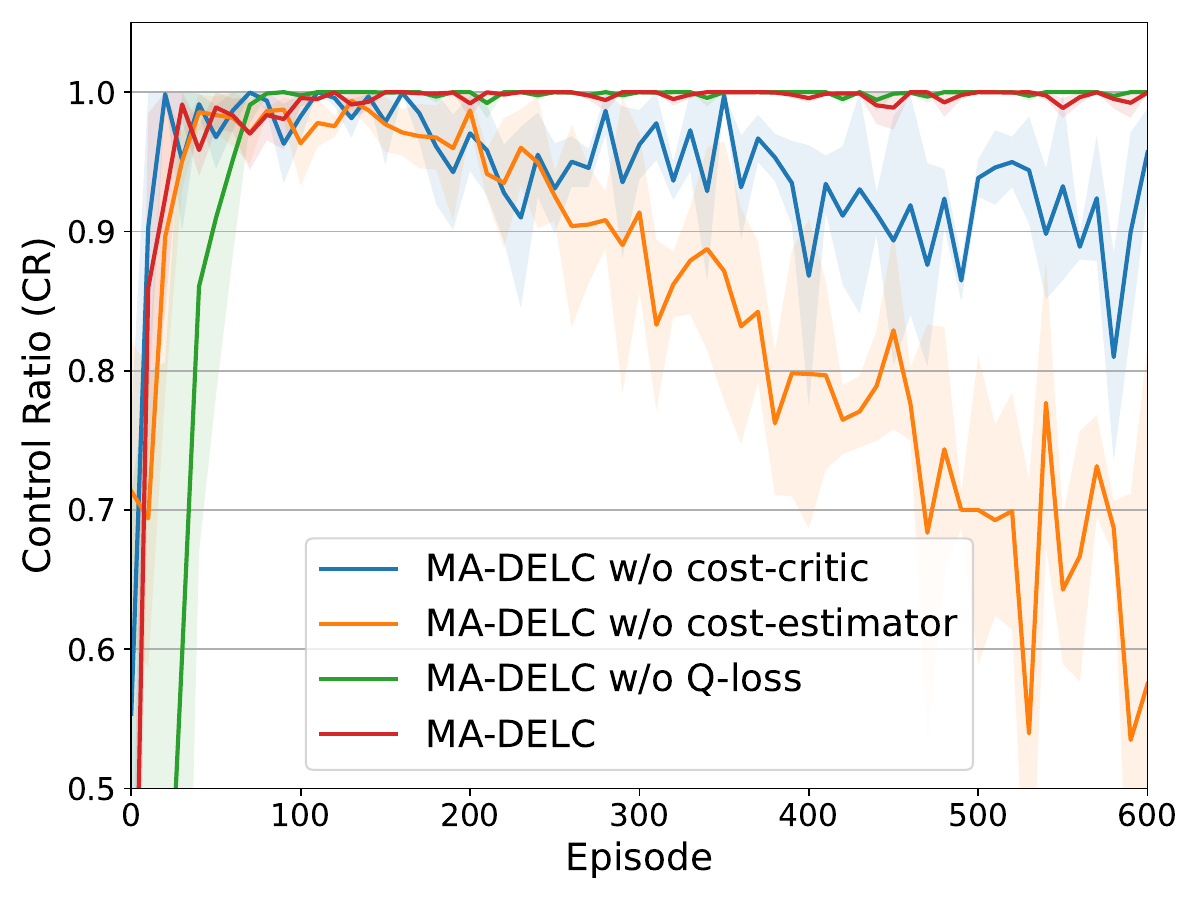}}
 	\vspace{3pt}
 	\centerline{\includegraphics[width=\textwidth]{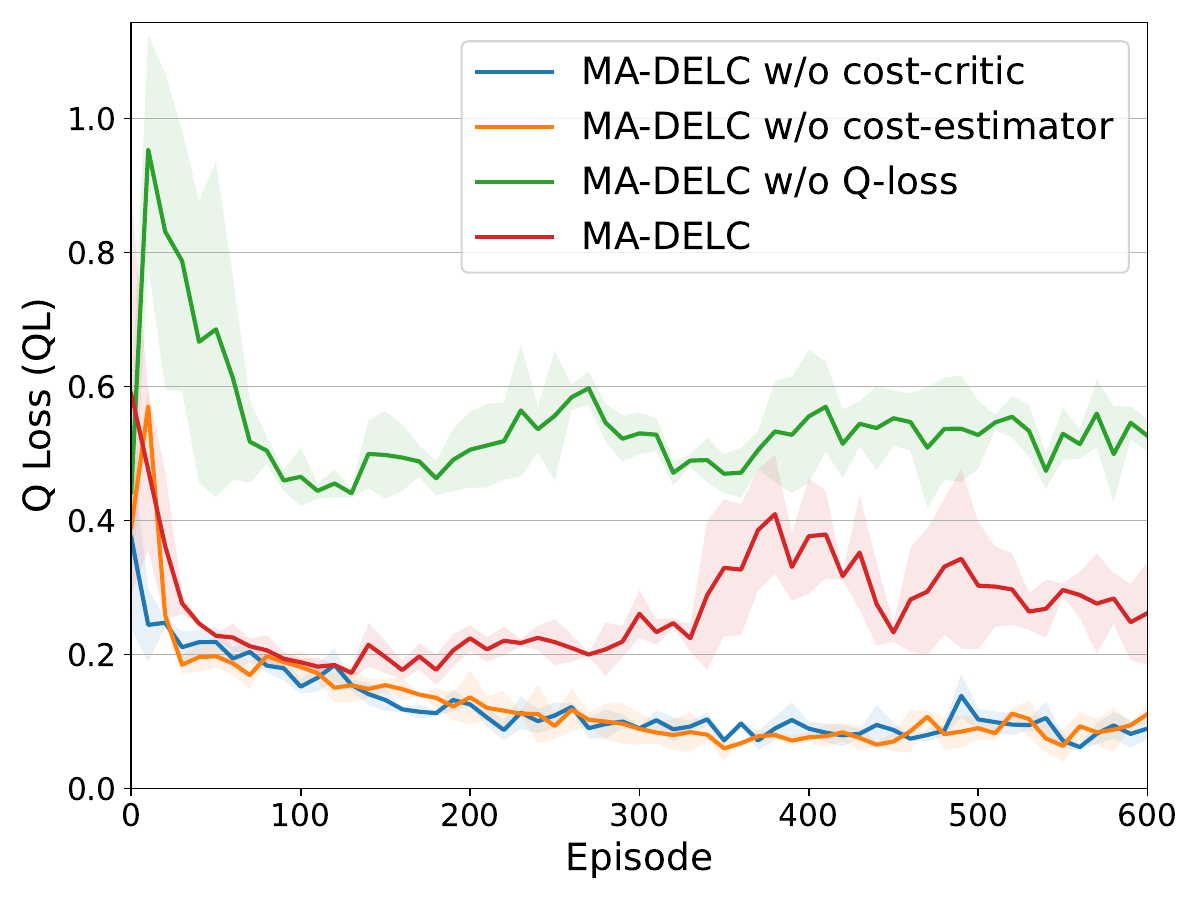}}
 	\vspace{3pt}
 	\centerline{(a) 33-bus scenario}
 \end{minipage}\hfill
 \begin{minipage}{0.3\linewidth}
	\vspace{3pt}
 	\centerline{\includegraphics[width=\textwidth]{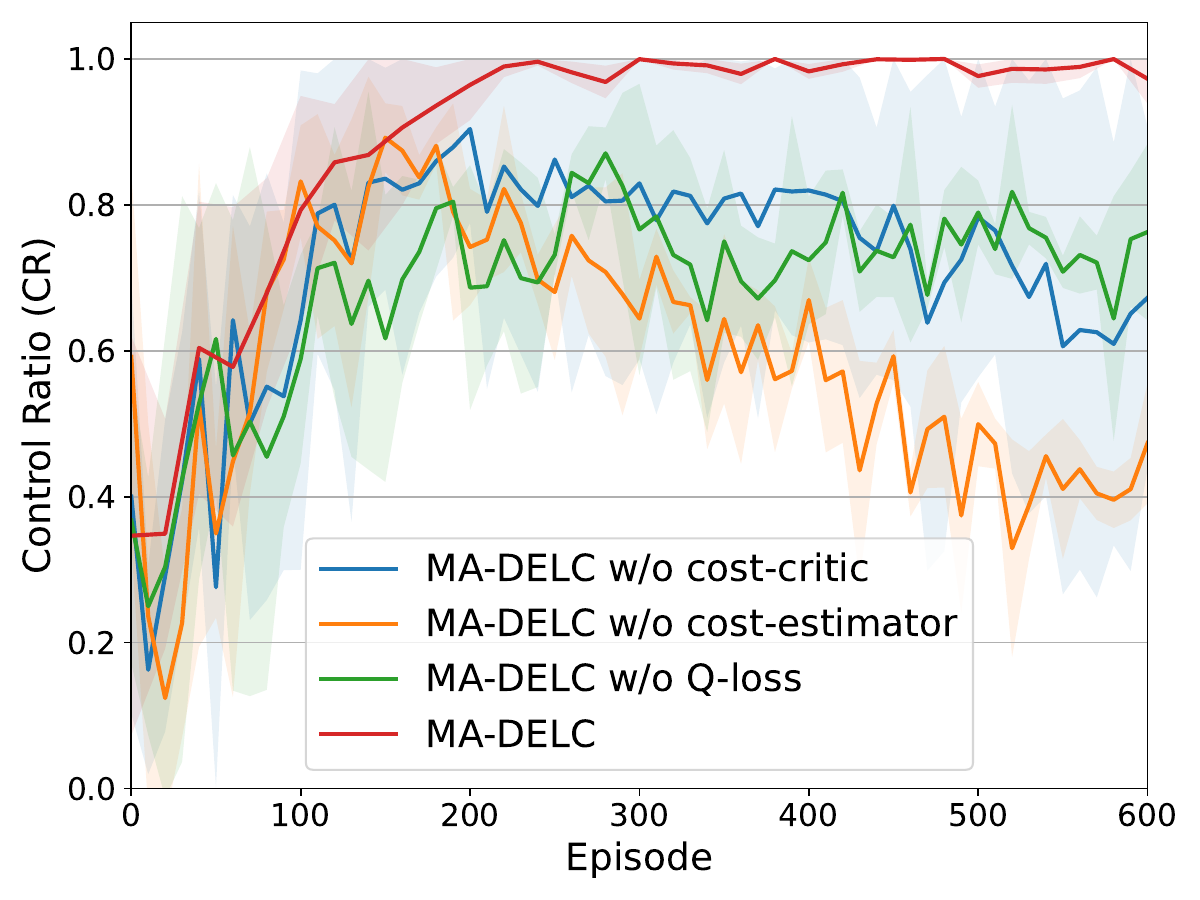}}
 	\vspace{3pt}
 	\centerline{\includegraphics[width=\textwidth]{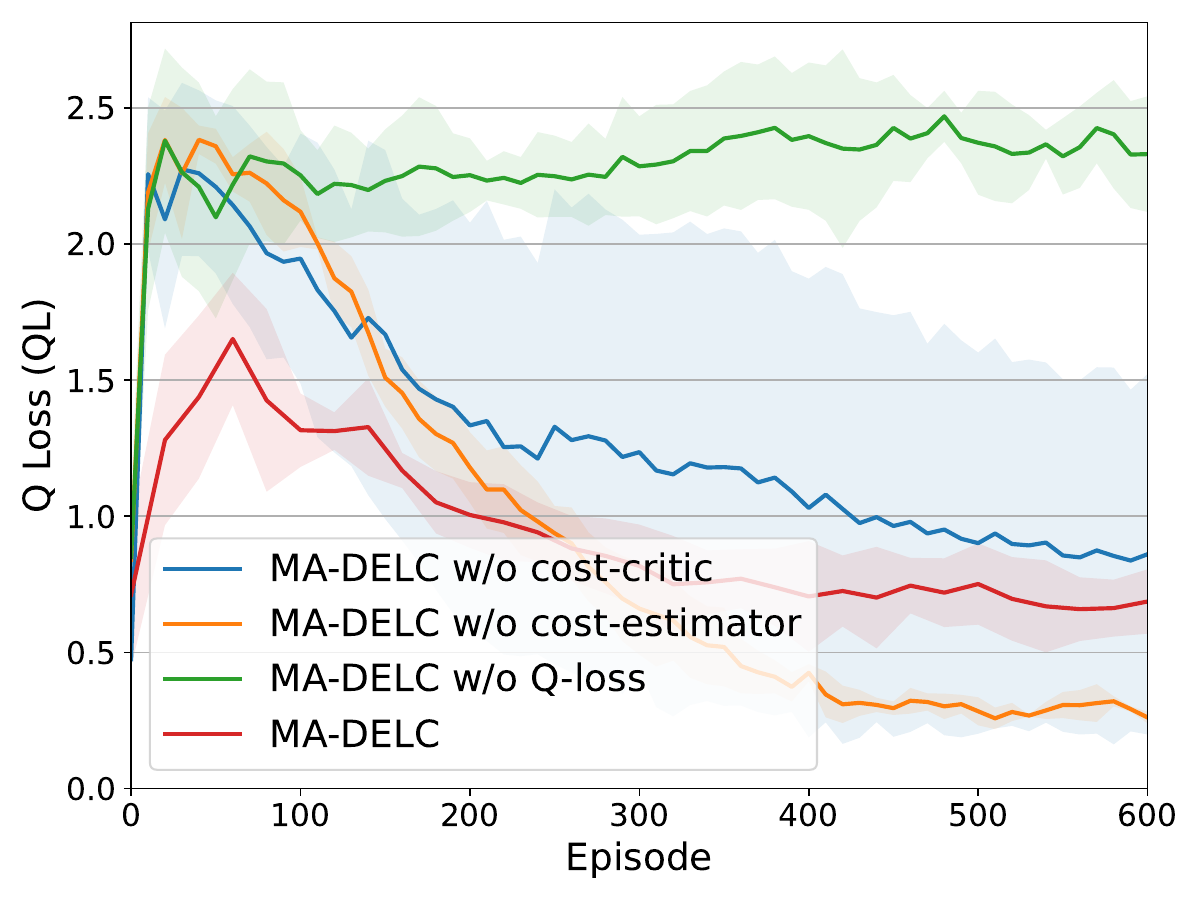}}
 	\vspace{3pt}
	\centerline{(b) 141-bus scenario}
\end{minipage}\hfill
\begin{minipage}{0.3\linewidth}
	\vspace{3pt}
 	\centerline{\includegraphics[width=\textwidth]{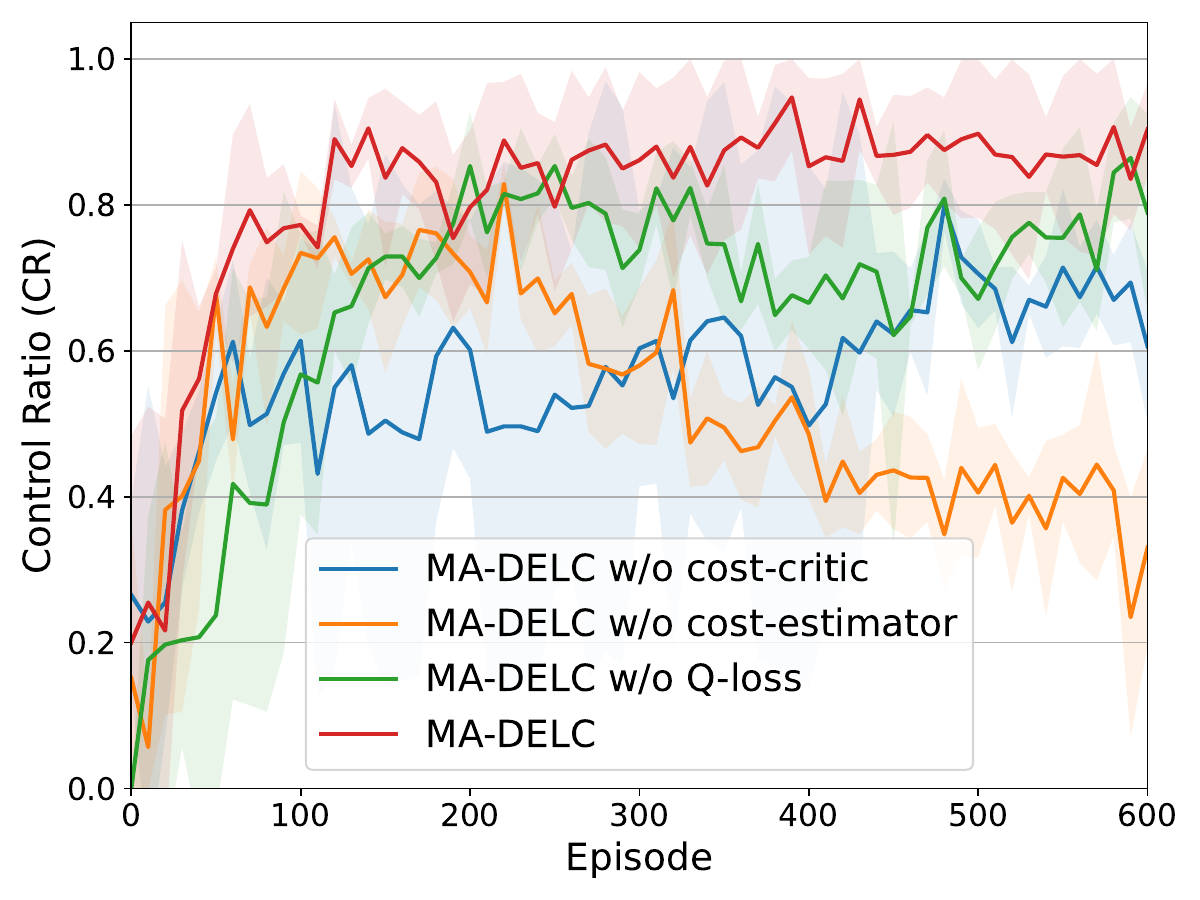}}
 	\vspace{3pt}
 	\centerline{\includegraphics[width=\textwidth]{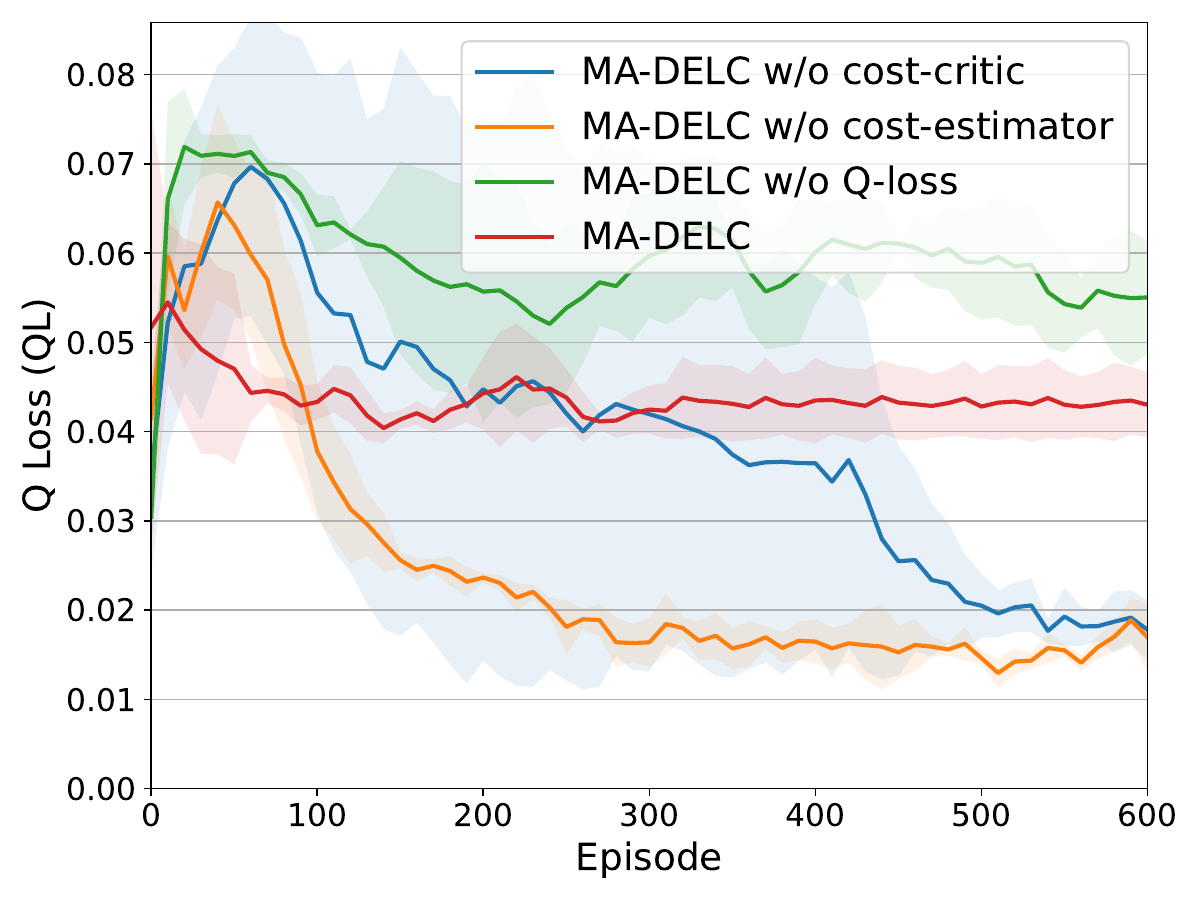}}
 	\vspace{3pt}
	\centerline{(c) 322-bus scenario}
 \end{minipage}
\caption{Control Ratio (CR) and Q Loss (QL) results of the ablation studies (CR: higher is better, QL: lower is better).
}
\label{fig:ablation}
\end{figure*}

\subsection{Experiment Setups}\label{expsettings}

\paragraph{Environment.}
We conduct experiments based on the Multi-Agent Power Distribution Networks (MAPDN) environment \cite{nips2021MARLavc}. 
The MAPDN environment serves as a suitable framework for distributed active voltage control, readily accommodating the application of MARL algorithms. 
This environment presents 3 scenarios: the 33-bus, the 141-bus and the 322-bus scenarios, containing 6, 22 and 38 agents, 32, 84 and 337 loads, respectively.
The load and PV data is sourced from real-world datasets, lending authenticity to the simulations. The topological structures of the 141-bus and 322-bus networks are visualized in Figure~\ref{fig:network}.

To assess the performance of the MARL algorithms in these scenarios, we employ two key evaluation metrics:
\begin{itemize}
  \item \textbf{Controllable Ratio (CR)}: It calculates the ratio of the steps in an episode where all buses' voltage is under control within the safety range (i.e. $0.95\le v_i\le 1.05, \forall i \in V\backslash \{0\}$). 
  \item \textbf{Q Loss (QL)}: It calculates the mean reactive power generations by agents per time step. This metric is chosen as a proxy for power loss (PL), which is often challenging to obtain for the entire network. QL serves as an approximation of the power loss attributable to PVs (agents).
\end{itemize}
These metrics together enable a comprehensive assessment of the algorithms' capabilities in terms of voltage control and power loss. 
In our evaluation, our focus is to identify algorithms that achieve high CR while maintaining low QL, with CR being the primary concern.

\paragraph{Baseline Methods.}
According to the experimental results of \cite{nips2021MARLavc}, COMA \cite{COMA}, MADDPG \cite{MADDPG} and MATD3 \cite{MATD3} surpass other MARL algorithms in the MAPDN environment. Therefore, we use them as our baselines. We provide brief descriptions of each of these baselines below:
\begin{itemize}
  \item \textbf{COMA} \cite{COMA}: the method incorporates counterfactual reasoning to address credit assignment problems in cooperative multi-agent settings. 
  \item \textbf{MADDPG} \cite{MADDPG}: the method employs actor-critic architectures and centralized training with decentralized execution to enhance the coordination of agents. In the architecture, each agent has a centralized critic and decentralized actor. 
  \item \textbf{MATD3} \cite{MATD3}: the method combines the advantages of actor-critic methods with twin critics and target policy smoothing, making it suitable for complex multi-agent environments. 
\end{itemize}

The reward of these non-Lagrangian baselines follows the settings in \cite{nips2021MARLavc}, defined as $r=-\frac{1}{|V|}\sum_{i\in V}l_v(v_i)-\beta\cdot l_q(q^{PV})$. $l_v(\cdot)$ is a voltage barrier function to measure voltage deviation from $1.0p.u.$; 
$\beta$ is a hyper-parameter to trade-off between satisfying safety constraints and reducing power loss; 
$l_q(q^{PV})$ is the reactive power generation loss of PVs. 

\begin{figure*}[t]
 \begin{minipage}{0.3\linewidth}
 	\vspace{3pt}
 	\centerline{\includegraphics[width=\textwidth]{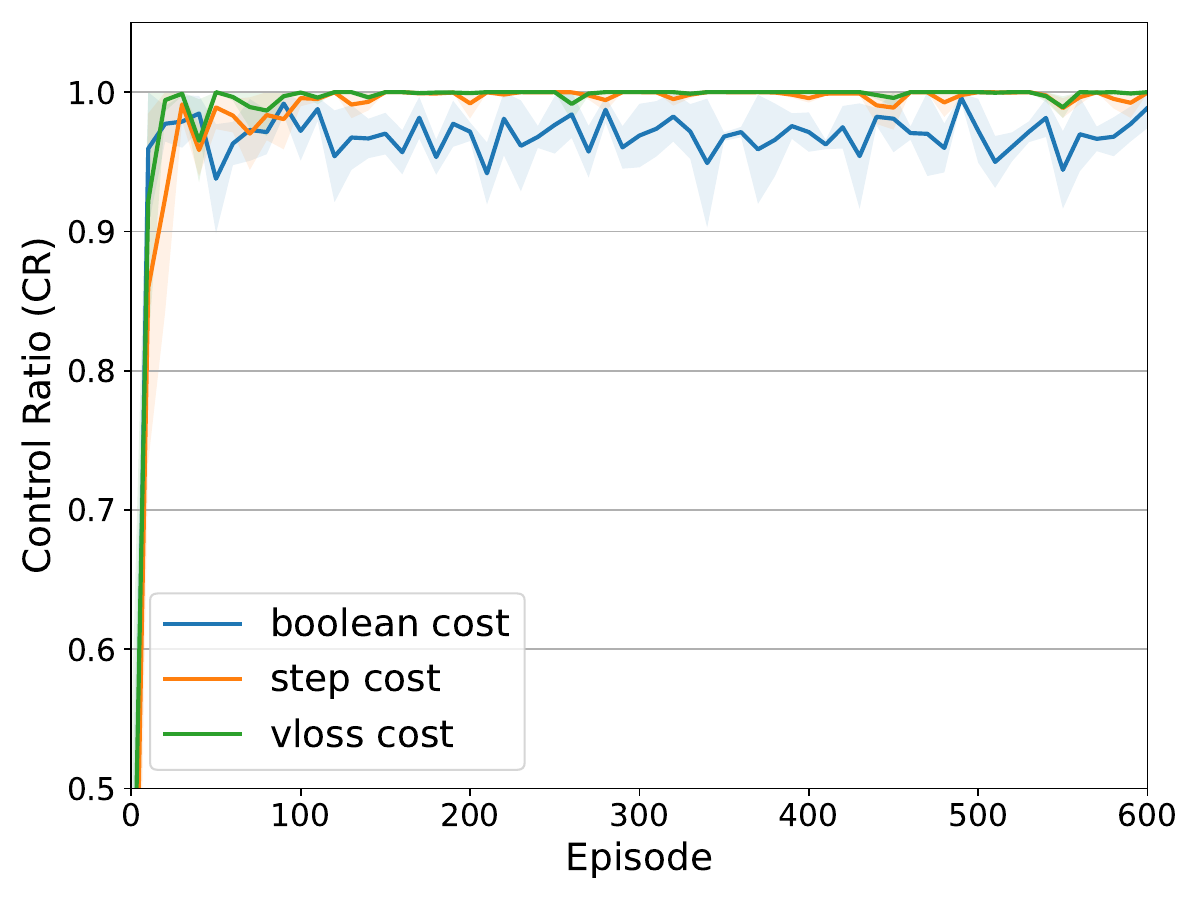}}
 	\vspace{3pt}
 	\centerline{\includegraphics[width=\textwidth]{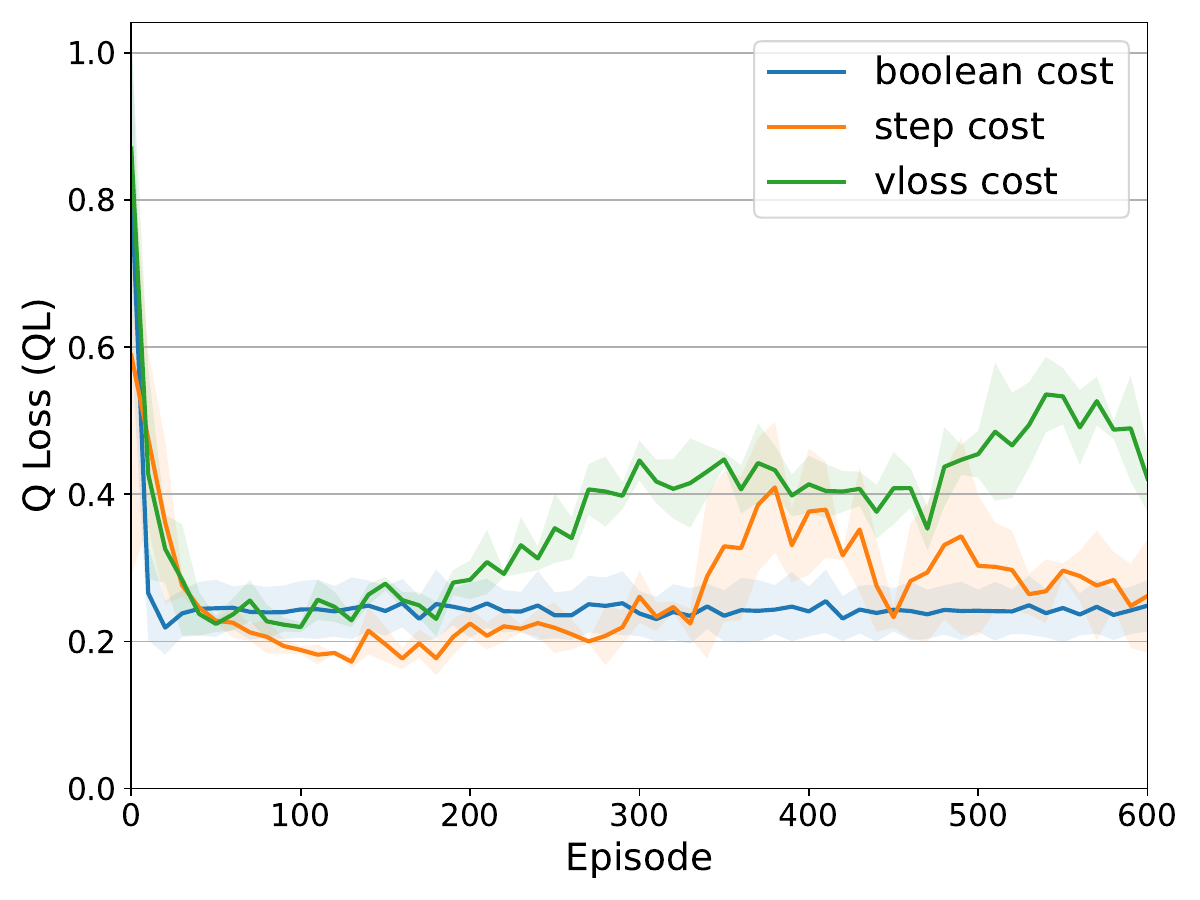}}
 	\vspace{3pt}
 	\centerline{(a) 33-bus scenario}
 \end{minipage}\hfill
 \begin{minipage}{0.3\linewidth}
	\vspace{3pt}
 	\centerline{\includegraphics[width=\textwidth]{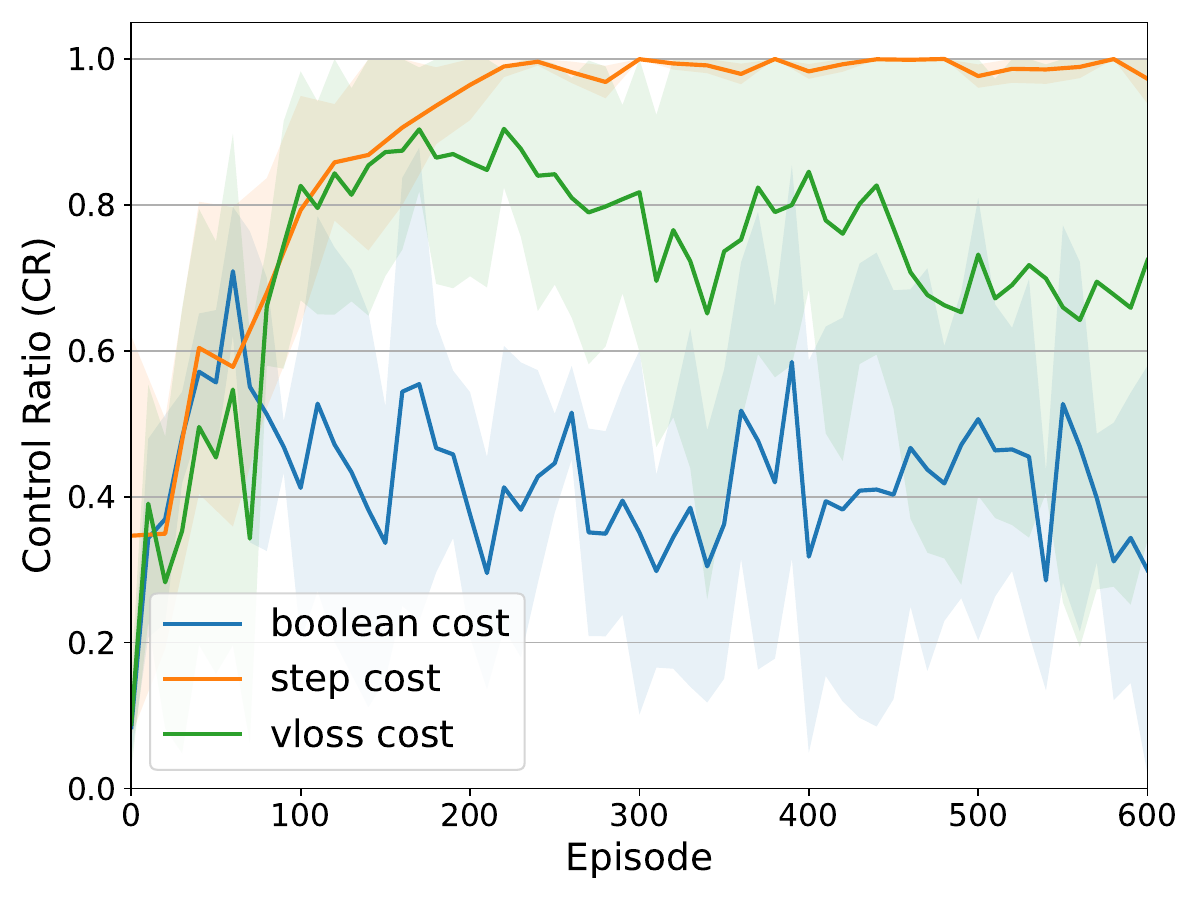}}
 	\vspace{3pt}
 	\centerline{\includegraphics[width=\textwidth]{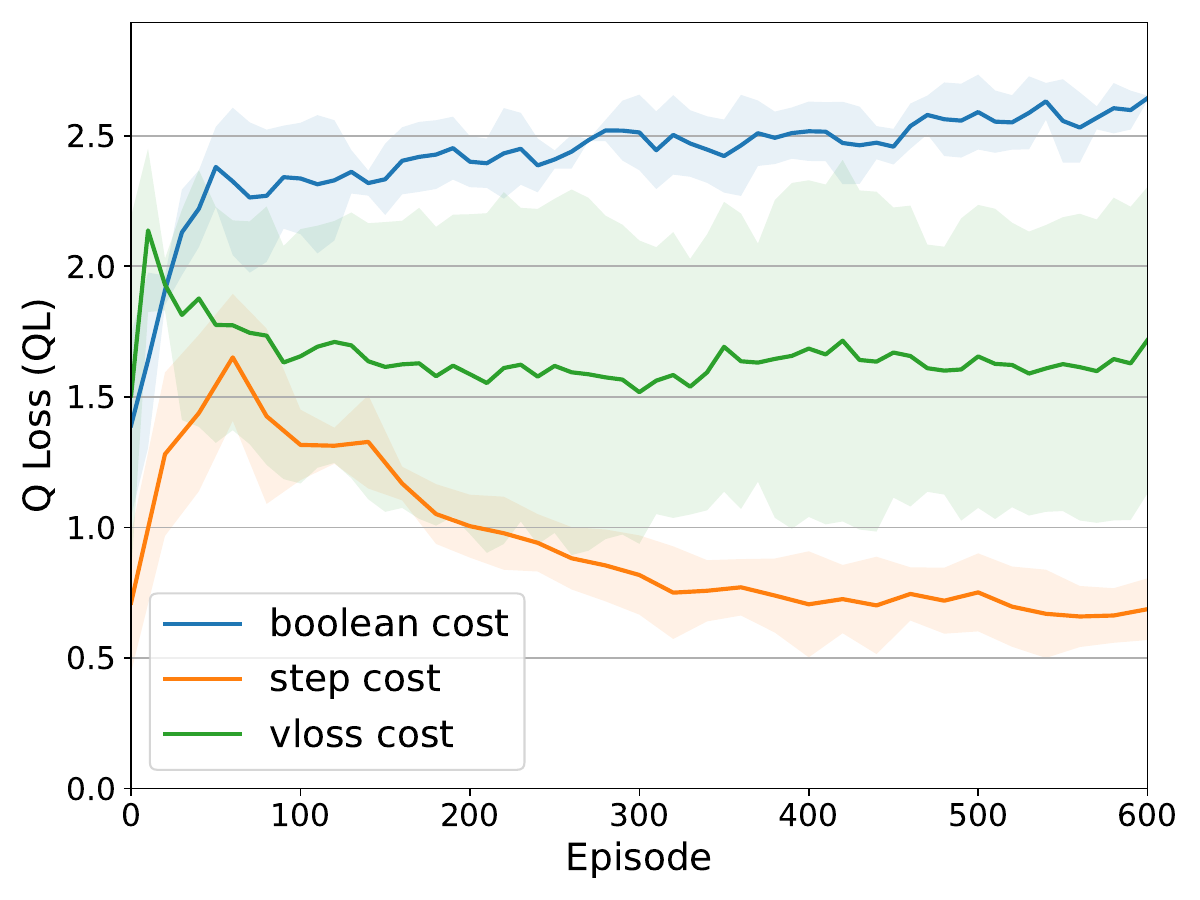}}
 	\vspace{3pt}
	\centerline{(b) 141-bus scenario}
\end{minipage}\hfill
\begin{minipage}{0.3\linewidth}
	\vspace{3pt}
 	\centerline{\includegraphics[width=\textwidth]{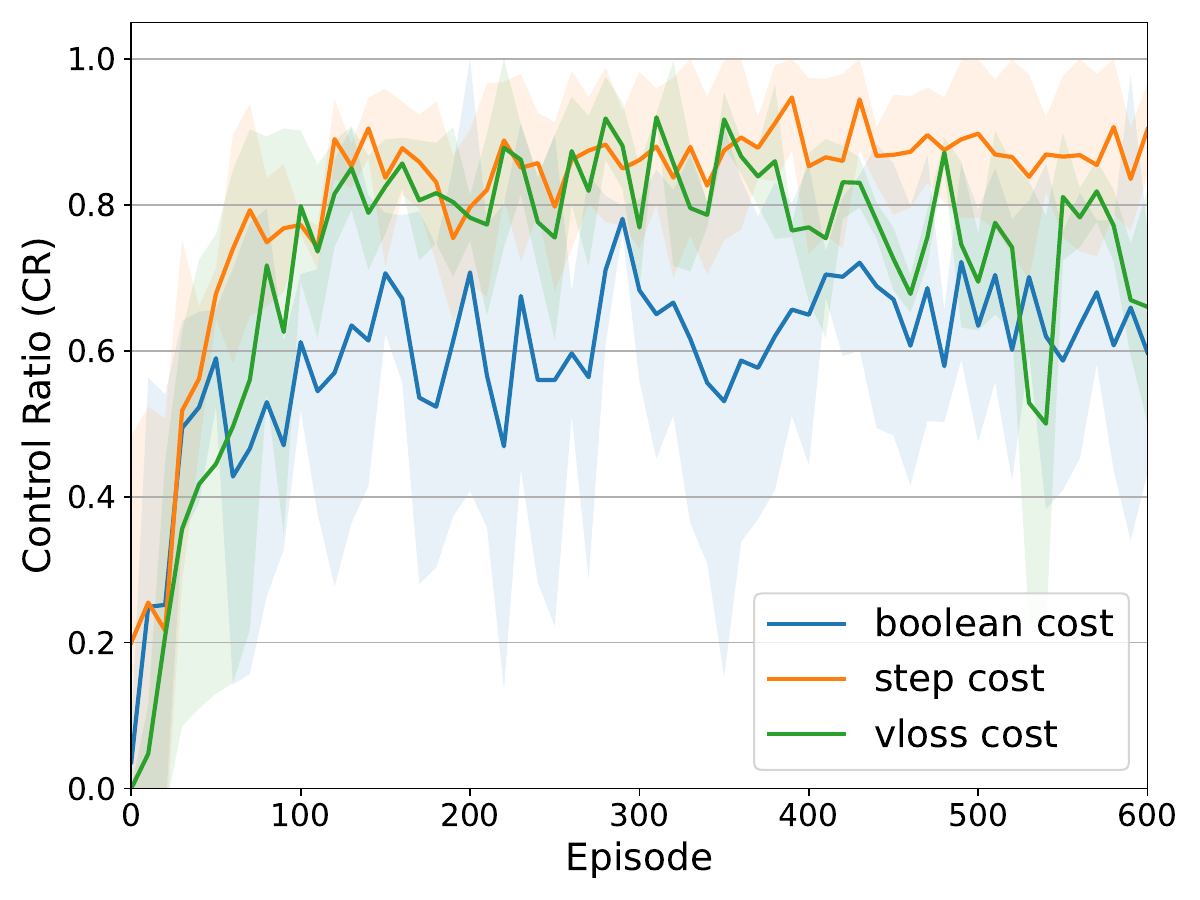}}
 	\vspace{3pt}
 	\centerline{\includegraphics[width=\textwidth]{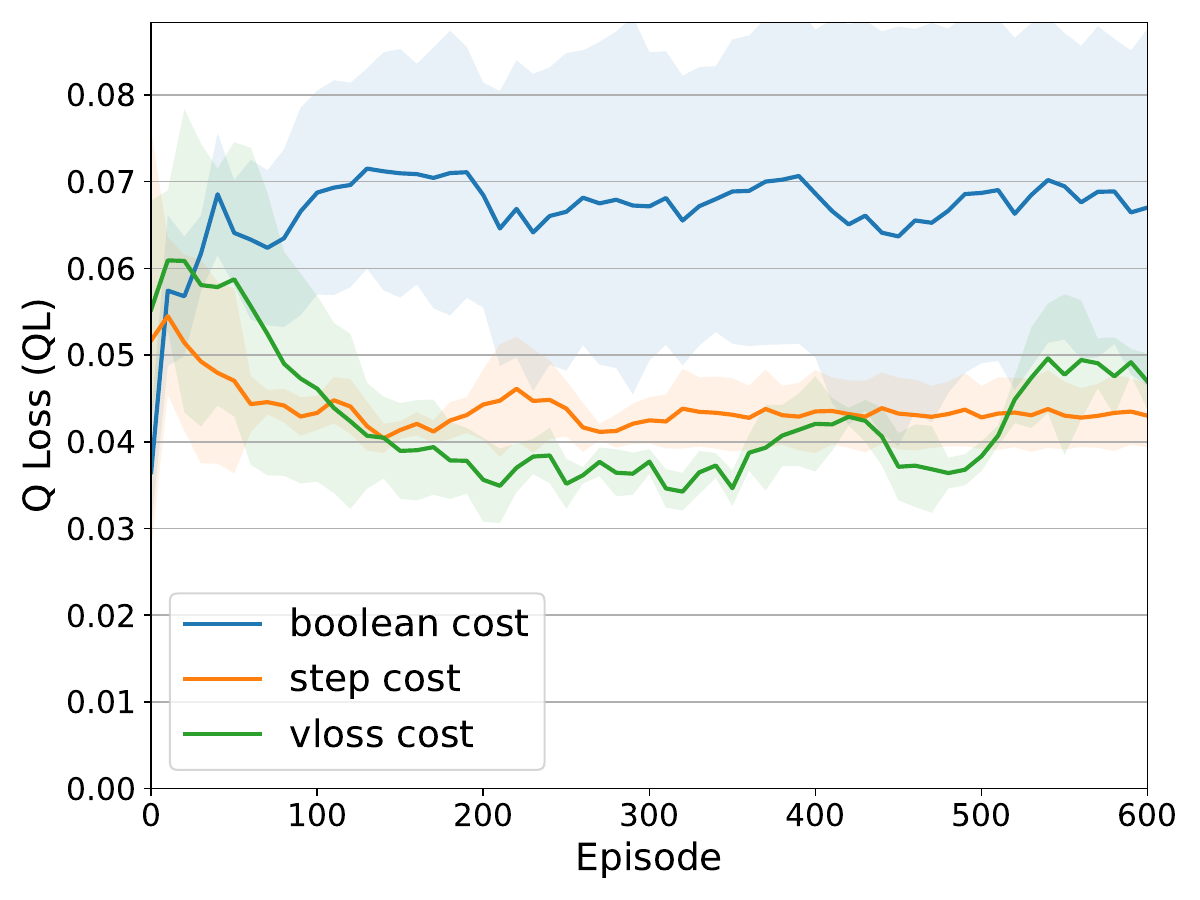}}
 	\vspace{3pt}
	\centerline{(c) 322-bus scenario}
 \end{minipage}
\caption{Control Ratio (CR) and Q Loss (QL) results of the different cost functions (CR: higher is better, QL: lower is better).}
\label{fig:costcomparison}
\end{figure*}

\subsection{Results}
The median results of CR and QL of various algorithms are shown in Figure~\ref{fig:result}, and the shaded areas in the charts represent the standard deviations of runs with different random seeds. 

As illustrated in the figure, our proposed approach (MA-DELC) outperforms the other baseline methods in CR in all of the 3 scenarios. Specifically, MADDPG achieves a CR of approximately 0.92 in the 33-bus scenario and around 0.9 in the 141-bus scenario. COMA's and MATD3's performance is comparable to MADDPG regarding CR in the 33-bus scenario but deteriorates in the 141-bus scenario. 
Remarkably, our proposed MA-DELC approach attains a stable CR close to 1.0 in these two scenarios. The QL of MA-DELC is slightly higher than baselines in the 33-bus scenario, but we consider it acceptable for maintaining an outstandingly stable CR. 
This achievement underscores the efficacy of the Lagrangian constraint method in consistently satisfying the constraints while optimizing the objective function. 

In the more complex 322-bus scenario, all baselines exhibits unstable
CR and low QL of 0.03-0.05. Among the baselines, MADDPG maintains relatively higher CR (around 0.75) and lowest QL (around 0.03).  MA-DELC outperforms other baselines in terms of CR, and achieves a QL close to COMA. 
The results of CR of MA-DELC and other baselines all exhibit high variance. We assume that in the 322-bus scenario, the intricate topological structure of the power grid demands more sophisticated agent coordination, and achieving a CR of 1.0 may be particularly challenging or even infeasible. Additionally, according to the results in \citet{AAMAS-safetylayer}, the safety layer method reaches CR of 0.95 , 0.95, 0.85 in the 3 scenarios, while our method reaches a higher CR of 1.0, 1.0 and 0.9 respectively apart from reducing power loss. Overall, our results highlight the promising performance of MA-DELC, especially in scenarios with high complexity and stringent constraints, when applying to real-world.

\subsection{Ablation Studies}
We conduct ablation experiments to systematically assess the impact of each component within our proposed method, as well as the effects of different cost functions.

\paragraph{Effect of Each Component.}
The results of our ablation experiments are depicted in Figure~\ref{fig:ablation}. 
MA-DELC w/o cost-critic uses the single step cost estimator in training the actor network, and MA-DELC w/o cost-estimator uses the cost-critic to update the Lagrange multiplier$\alpha$. MA-DELC w/o Q-loss uses only the cost function in training.

MA-DELC w/o cost-critic exhibits lower CR in all of the three scenarios, which shows that long-term planning is necessary for satisfying the constraints in active voltage control problem. On the other hand, MA-DELC w/o cost-estimator can achieve high CR in the middle of the training, while fails to maintain it due to optimistic overestimation of cost-critic. These results validate the necessity of the proposed double-estimation of Lagrangian constraint approach. 

MA-DELC w/o Q-loss exhibits clearly higher QL, which verifies that MA-DELC can fulfill the constraints and optimize the objective concurrently. 
Notably, MA-DELC has higher CR than MA-DELC w/o Q-loss in the 141-bus and 322-bus scenario. We assume that it is because maintaining the constraint fulfillment and optimizing the objective do not always present the opposite gradient, and Q-loss provides agents with useful information about the environment as well.

\paragraph{ Comparisons for Cost Functions.} \label{subsection:costexp}
As shown in Figure~\ref{fig:costcomparison}, we explore alternative cost functions and provide the comparative performance in terms of CR and QL. 
In general, the step cost function stands out as the most effective choice, outperforming the other cost functions in both CR and QL. 
These results echo our initial argument that cost functions with more information facilitate exploration for the agents.

When employing boolean cost functions, MA-DELC manages to learn policies achieving a CR of 1.0 in the 33-bus scenario, but fails in the 141-bus and 322-bus scenarios. 
This discrepancy could be attributed to the limited informativeness of the boolean cost function, which may not provide sufficient guidance for the agents in the complex large scale scenario. 
On the other hand, MA-DELC with the v-loss cost function exhibits unsatisfactory CR in the 141-bus scenario, but is close to the step cost function in the 322-bus scenario. 
The v-loss cost function, while offering more information, seems to blur the boundary of the constraint fulfillment. 
In the 322-bus scenario, the v-loss cost function provides additional information that can be more helpful, as the constraints are rarely fully met, leading to better performance.

These insights into the behavior of different cost functions help us better understand their impact on the learning process and the ability of the algorithm to satisfy constraints effectively. 
Ultimately, the choice of cost function should be considered based on the specific characteristics and requirements of the power grid scenario.

\section{Conclusion} \label{conclusion}

In this paper, we focus on the active voltage control problem in power distribution networks installed with PVs. 
We propose MA-DELC, an innovative constrained MARL algorithm, with the double estimation framework and an adaptive Lagrange multiplier, to strike a balance between minimizing power loss and ensuring safety compliance. 
We conducted extensive experiments and comprehensive ablation studies within real-world-scale scenarios provided by the MAPDN dataset. 
Experimental results demonstrate that MA-DELC with appropriate cost functions outperforms other baseline approaches. 
Although motivated by the AVC problem, our method is model-free and does not need any domain-specific information about the physical system. 
In the future, we consider testing the algorithm in more complicated settings with real-world power systems, and the applying our method to other real-world safety-constrained problems.

\clearpage

\section*{Acknowledgments}
This work was supported in part by the Major Research Plan
of the National Natural Science Foundation of China (Grant No.
92048301) and Anhui Provincial Natural Science Foundation (Grant
No. 2208085MF172).

\bibliographystyle{named}
\bibliography{main}

\clearpage

\appendix
% 简介各部分
\noindent Due to limited space of the submitted paper, we supply the following materials for better understanding of our paper. 
Notably, we provide more experimental results with additional metrics, showing that our method does not only achieve higher controllable ratio in time steps, but also significantly reduces the voltage deviation of the power network. In all the tested scenarios, we can see that the percentage of the node with voltage out of control is almost 0, and the rates of maximum high and low voltages are almost 0 (see {\bf Figure \ref{fig:result_con}} for more details). This is critical to the safety of the power network because voltages out of range may damage the devices and loads connected to the power line. Moreover, we maintain similar power loss comparing with the other baselines (see {\bf Figure \ref{fig:result_obj}} for more details). Those additional experimental results further demonstrate the merits of our method by explicitly modeling the constrains of active voltage control. 

The appendix are organized as follows. In Appendix A, we explain the physical dynamics of the power grid, and give a formal definition of the active voltage control problem. 
Then in Appendix B, we provide more detailed information about the environment, dataset, algorithm hyper-parameters, and running settings.
Finally in Appendix C, we provide more experimental results of additional metrics to further confirm the effectiveness of the proposed method.

\section{The Constrained Optimization Model for the Active Voltage Control Problem}

The active voltage control problem can be modeled as a {\em constrained optimization}, where: 1) the constraints are voltage safety and the power flow rules, and 2) the optimization objective is the power loss.

Recall the graph modeling of the power distribution network. We model the power distribution network as a tree graph $\mathcal{G}=(V, E)$, where $V=\{0,1,\dots,|V|\}$ 
represents the set of nodes (buses), and $E=\{(i,j)|i, j\mbox{ connected}\}$ represents the set of edges (branches). Due to the acyclic nature of the power distribution network, we have $|E|=|V|-1$.
Nodes in the distribution network are divided into several zones based on their shortest path from the terminal to the main branch \cite{OptimalPower}. 
Here, Node 0 is connected to the main grid, which serves to balance the active and reactive power in the distribution network. Node 0 has a stable voltage magnitude, denoted as $v_0=1.0p.u.$. 
Each node may be connected to loads and/or PVs. 

Denote the complex admittance on branch $(i,j)$ as $y_{ij}=g_{ij}-\mathbf{i}b_{ij}$, where $g_{ij}, b_{ij}$ is the conductance and susceptance on branch $(i,j)$ respectively, and $\mathbf{i}$ is the imaginary unit. 
Moreover, the complex impedance on branch $(i,j)$ is denoted as $z_{ij}=\frac{1}{y_{ij}}=r_{ij}+\mathbf{i}x_{ij}$, where $r_{ij}, x_{ij}$ is the resistance and reactance on branch $(i,j)$ respectively. 

The power loss optimization object, or known as the line loss \cite{9805763}, is defined as below: 
\begin{equation}
    P_{loss} = \Sigma _{(i,j)\in E}r_{ij}|I_{ij}|^2
\end{equation}
where $r_{ij}$ is the resistance on branch $(i,j)$, and $|I_{ij}|$ is the current magnitude from bus $i$ to $j$. 

We denote the complex power injection of node $j$ as $s_j = p_j + \mathbf{i}q_j$. Let $p_i^{PV}, q_i^{PV}$ be the active and reactive power generation of PV in node $i$ respectively, and if node $i$ is not installed with PV, they are set to 0. 
Let $p_i^L, q_i^L$ be the active and reactive power demand of loads in node $i$. Also, if bus $i$ does not have any load, these values are set to 0. Then $p_i =p_i^{PV}-p_i^L$, $q_i =q_i^{PV}-q_i^L$. In more details, the active and reactive power injection can be computed as follow:
\begin{equation}\label{eq:power_rule}
\begin{split}
	p_i & =p_i^{PV}-p_i^L \\
	& =v_i^2\sum_{j\in V_i}g_{ij}- 
        v_i\sum_{j\in V_i}v_{j}\left(g_{ij}\cos\theta_{ij}+b_{ij}\sin\theta_{ij}\right), \\ 
        & \quad\quad \forall i \in V\backslash \{0\}, \\[10pt]
    q_i & =q_i^{PV}-q_i^L\\
	& =-v_i^2\sum_{j\in V_i}b_{ij}+ 
        v_i\sum_{j\in V_i}v_j\left(g_{ij}\sin\theta_{ij}+b_{ij}\cos\theta_{ij}\right), \\
        & \quad\quad \forall i \in V\backslash \{0\}
\end{split}
\end{equation}
where $V_i=\{j |(i,j) \in E\}$ is the set of buses connected to bus $i$, and
$\theta_{ij}=\theta_i-\theta_j$ is the phase difference of voltage between bus $i$ and $j$. 

Besides, buses in the power network follow the Ohm's law, and satisfy the current balance and the power balance \cite{OptimalPower} as shown below:
\begin{equation}
\begin{split}
        \mbox{Ohm's law: } & I_{ij}=y_{ij}(v_i-v_j), \forall (i,j)\in E, \\
        \mbox{Current balance: } & I_{i} = \sum_{j\in V_i} I_{ij}, \forall i \in V,\\
        \mbox{Power balance: } & s_i = v_iI_i^H, \forall i \in V
\end{split}
\end{equation}
where the superscript $H$ of $I_i^H$ denotes hermitian.

In summary, the active voltage control problem considered in this paper can be formulated as below:
\begin{equation}\label{eq:complete_avc} 
\begin{split}
\min ~~ & P_{loss} = \sum_{(i,j)\in E}r_{ij}|I_{ij}|^2,\\
\mbox{s.t. } ~~ & 0.95 p.u. \le v_i \le 1.05 p.u., \forall i \in V\backslash \{0\}; \\
& v_0=1.0 p.u. ; \\
& y_{ij}=g_{ij}-\mathbf{i}b_{ij}, \forall (i,j)\in E; \\
& z_{ij}=\frac{1}{y_{ij}}, \forall (i,j)\in E; \\
& z_{ij}=r_{ij}+\mathbf{i}x_{ij}, \forall (i,j)\in E; \\
& I_{ij}=y_{ij}(v_i-v_j), \forall (i,j)\in E; \\
& I_{i} = \sum_{j\in V_i} I_{ij}, \forall i \in V;\\
& s_i = v_iI_i^H, \forall i \in V; \\
& s_i = p_i+\mathbf{i}q_i, \forall i \in V;\\
& p_i=v_i^2\sum_{j\in V_i}g_{ij}- 
v_i \!\! \sum_{j\in V_i}v_{j}\left(g_{ij}\cos\theta_{ij}+b_{ij}\sin\theta_{ij}\right),  \\
& \quad\quad \forall i \in V\backslash \{0\}; \\
& q_i=-v_i^2\sum_{j\in V_i}b_{ij}+ 
v_i \!\! \sum_{j\in V_i}v_{j}\left(g_{ij}\sin\theta_{ij}+b_{ij}\cos\theta_{ij}\right), \\
& \quad\quad \forall i \in V\backslash \{0\}; \\
\end{split}
\end{equation}

The constrained optimization problem defined above is based on instantaneous setting, 
whereas we address the RL problem that necessitates planning over a longer time horizon. 
In the RL setting, voltage phase change is calculated by the simulator, and the active and reactive power demand of loads is obtained from the dataset. The active power generation of PVs is obtained from the dataset, and the reactive power generation of PVs is decided by agents' action. The admittance and impedance of each branch is recorded in the network model of the dataset. 
The physical constraints in Equation~\ref{eq:complete_avc} must be satisfied in every step, and the simulator solves equations of the physical constraints using Newton-Raphson method to calculate bus voltage and branch current. The simulator for RL environment is built based on pandapower \cite{pandapower}. 
The actions of agents have a enduring influence on network dynamics, so long-term planning is necessary.

\section{More Details of the Experimental Settings}

The MAPDN environment \cite{nips2021MARLavc} contains 3 scenarios varying in scale, and details about the scale are shown in Table~\ref{tab:env}. 
The MAPDN environment also provides real-world datasets of loads and PV penetration for each scenario. 
The datasets contains 3 years of real-world data in 2012-2015, with the recording interval being 3 minutes. 
The dataset for each scenario contains a model of the power grid network, data of active and reactive power demand of loads, and data of active power generation of PVs. An overview of the dataset is recorded in Table 2.

\begin{table}[ht]
        \centering \renewcommand{\arraystretch}{1.2}
        \caption{The scale of networks in the MAPDN environment.}
        \begin{tabular}{l|ccc}
            \hline
            Scenario & Loads & PVs & Regions \\
            \hline
            33-bus  & 32  & 6  & 4   \\
            141-bus & 84  & 22 & 9     \\
            322-bus & 337 & 38 & 22 \\
            \hline
        \end{tabular}
        \label{tab:env}
    \end{table}
\begin{table}[ht]
        \centering \renewcommand{\arraystretch}{1.3}
        \caption{Overview of datasets in the MAPDN environment.}
        \begin{tabular}{l|ccc}
            \hline
            Scenario & $v_0$ & $p_{max}^L$ & $p_{max}^{PV}$ \\
            \hline
            33-bus  & 12.7 kV  & 3.5 MW  & 8.8 MW  \\
            141-bus & 12.5 kV  & 20 MW & 80 MW    \\
            322-bus & 20 kV & 1.5 MW & 3.8 MW \\
            \hline
        \end{tabular}
        \label{tab:dataset}
    \end{table}

\begin{figure*}[ht!]	
 \begin{minipage}{0.32\linewidth}
 	\vspace{3pt}
 	\centerline{\includegraphics[width=\textwidth]{fig/CR-33bus}}
 	\vspace{3pt}
 	\centerline{\includegraphics[width=\textwidth]{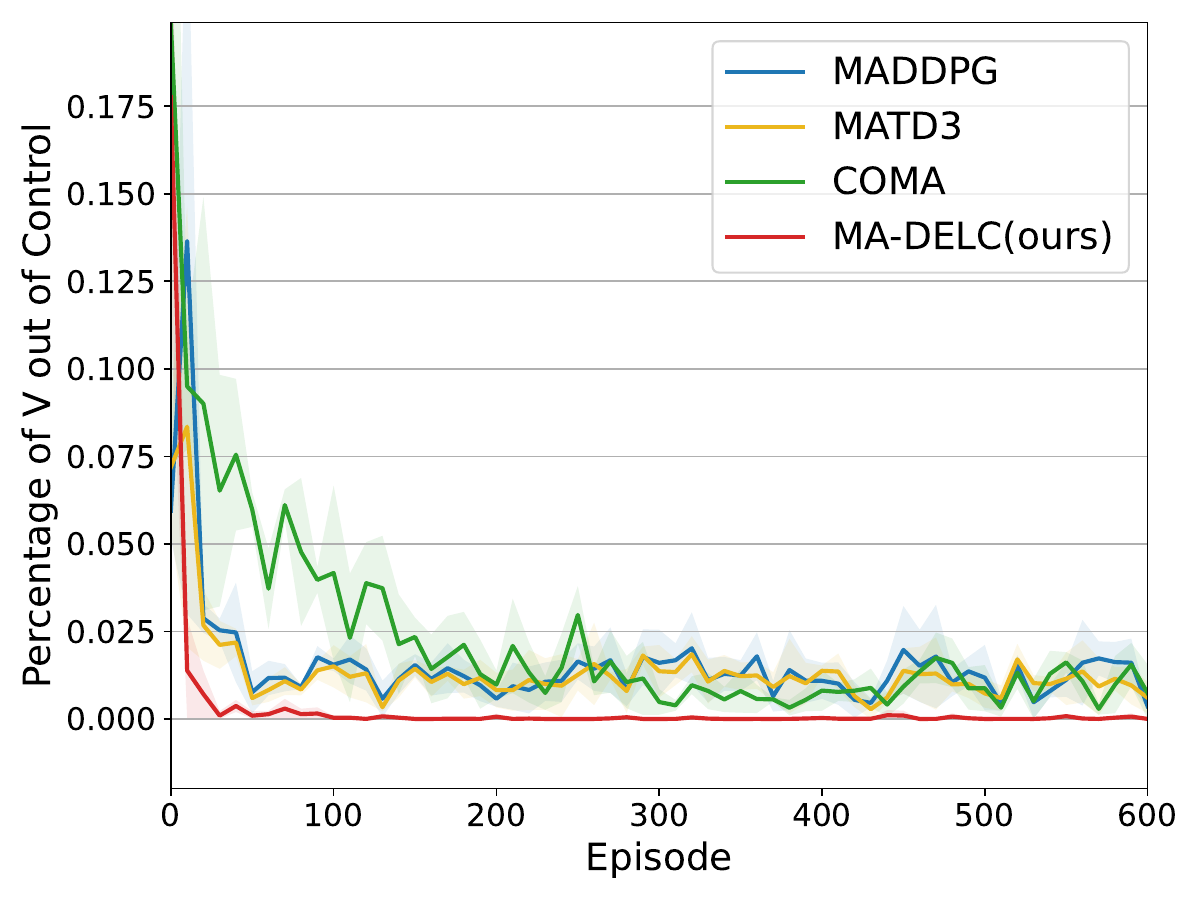}}
        \vspace{3pt}
 	\centerline{\includegraphics[width=\textwidth]{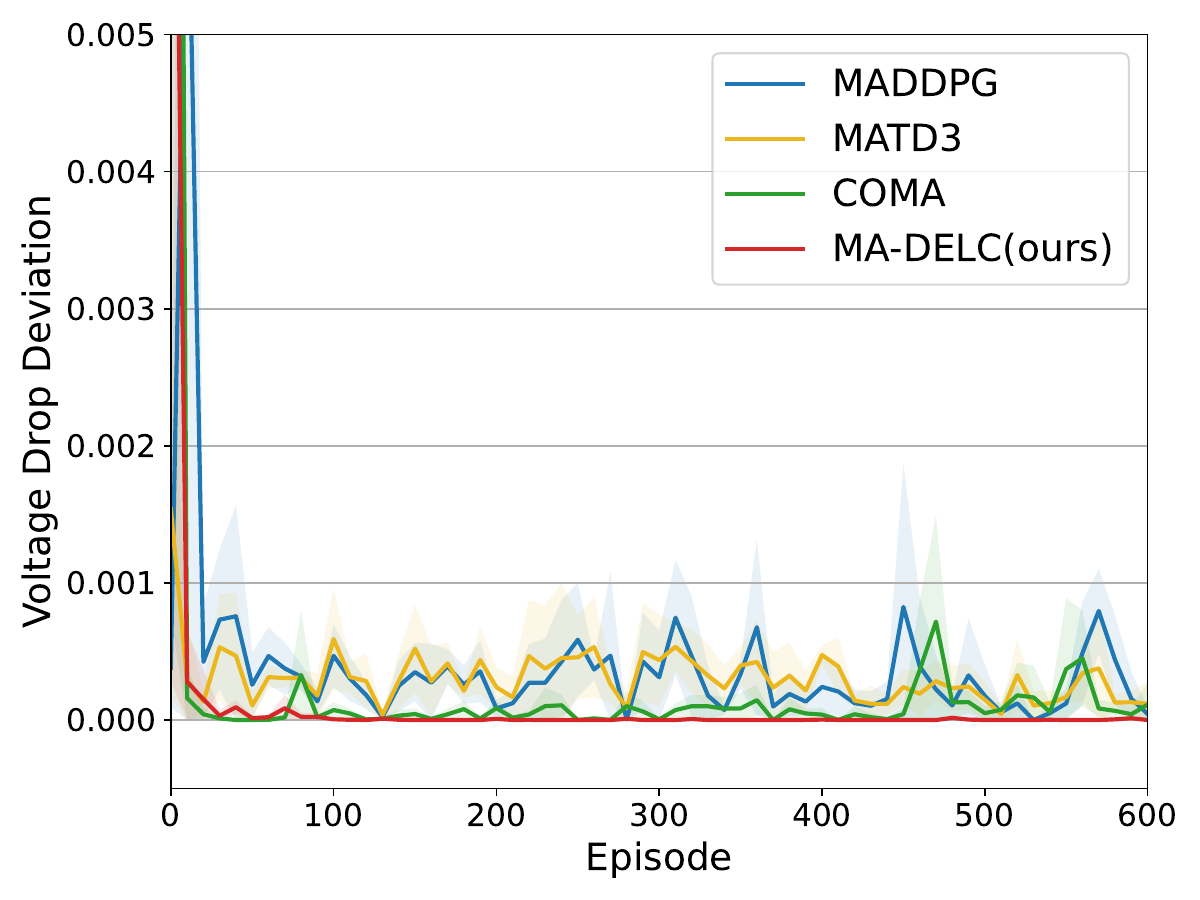}}
        \vspace{3pt}
 	\centerline{\includegraphics[width=\textwidth]{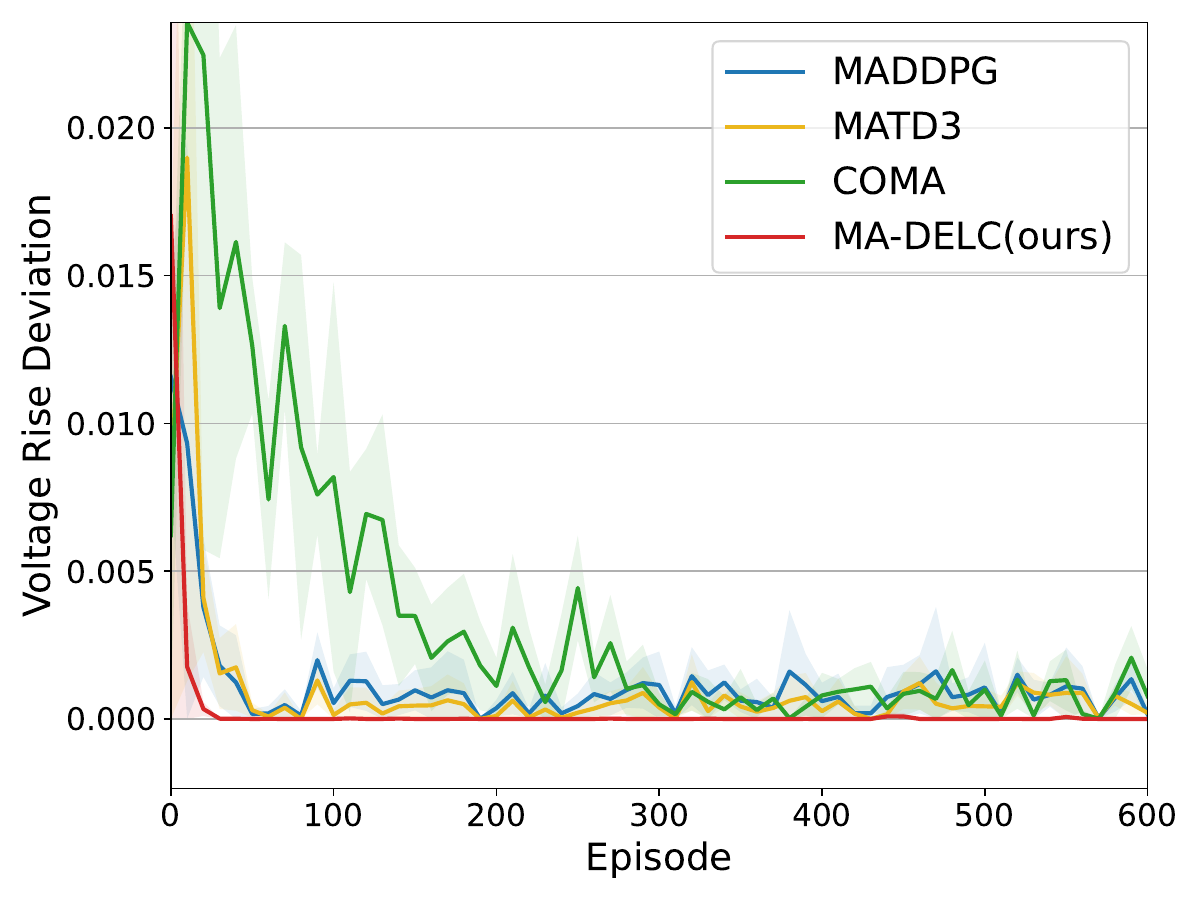}}
 	\vspace{3pt}
 	\centerline{(a) 33-bus scenario}
 \end{minipage}\hfill
 \begin{minipage}{0.32\linewidth}
	\vspace{3pt}
 	\centerline{\includegraphics[width=\textwidth]{fig/CR-141bus}}
 	\vspace{3pt}
 	\centerline{\includegraphics[width=\textwidth]{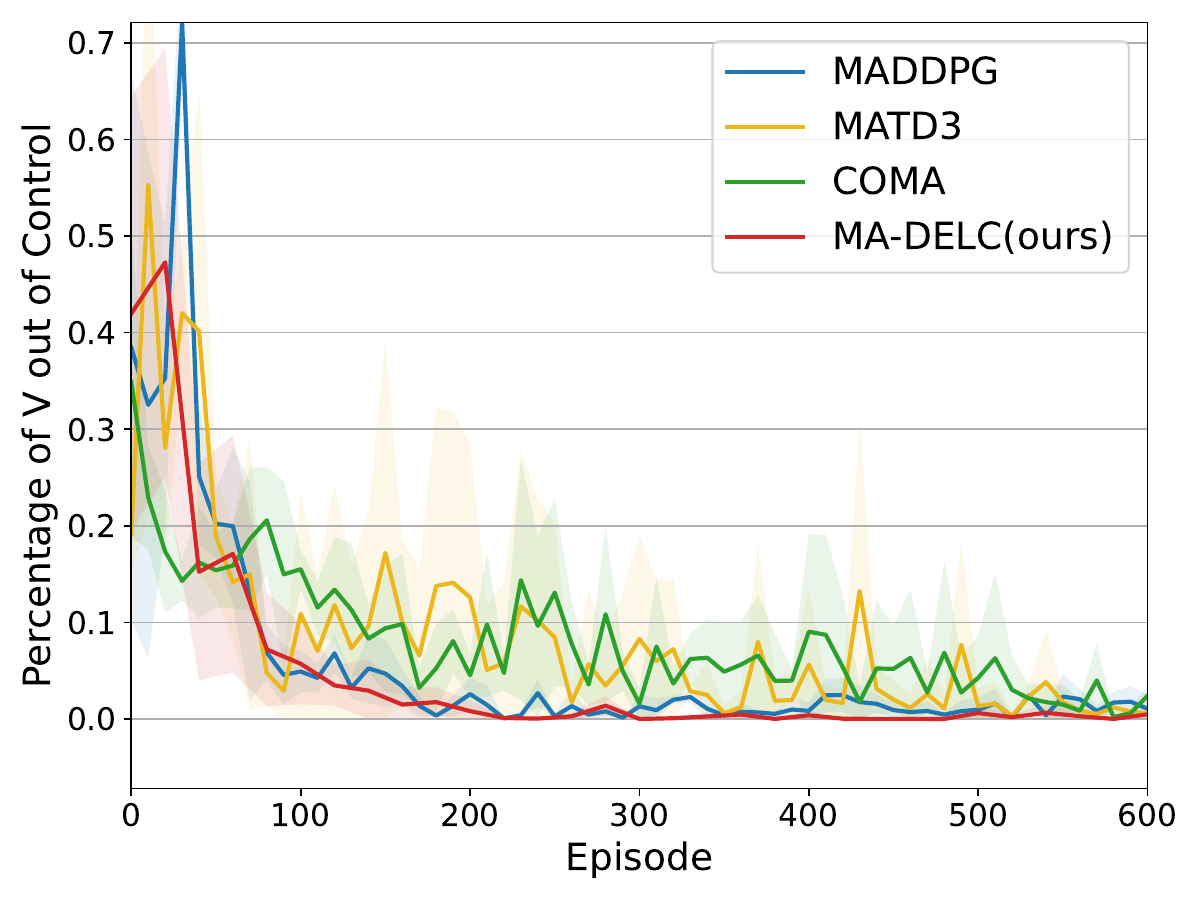}}
        \vspace{3pt}
 	\centerline{\includegraphics[width=\textwidth]{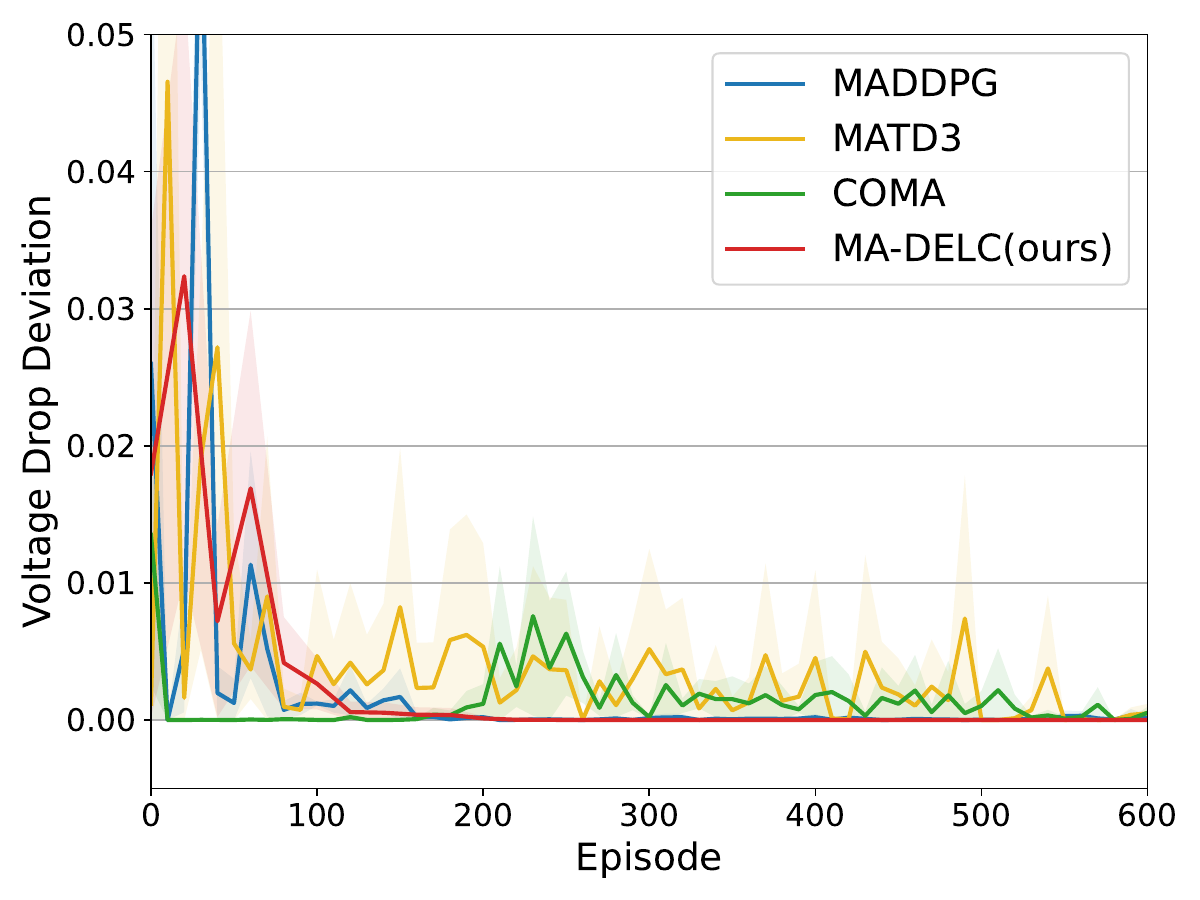}}
        \vspace{3pt}
 	\centerline{\includegraphics[width=\textwidth]{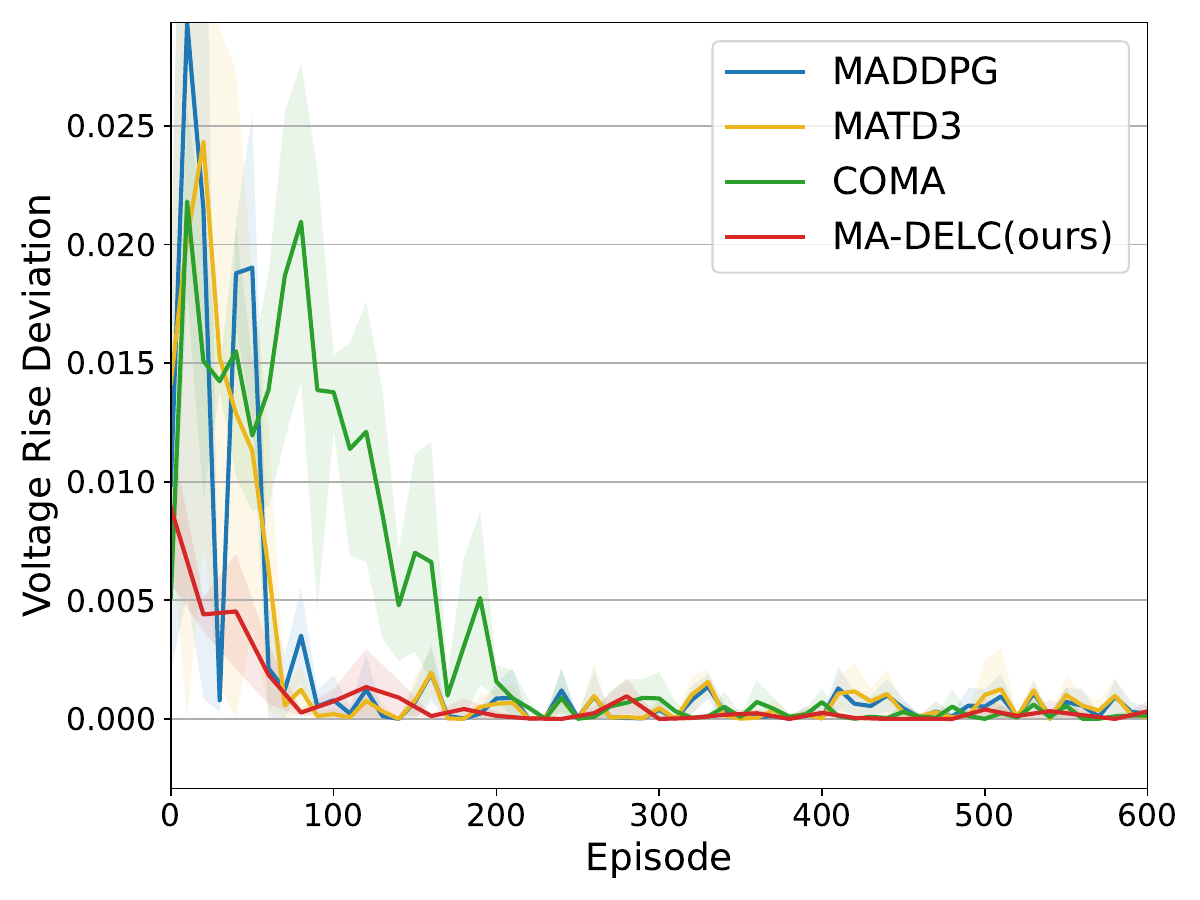}}
 	\vspace{3pt}
	\centerline{(b) 141-bus scenario}
\end{minipage}\hfill
\begin{minipage}{0.32\linewidth}
	\vspace{3pt}
 	\centerline{\includegraphics[width=\textwidth]{fig/CR-322bus}}
 	\vspace{3pt}
 	\centerline{\includegraphics[width=\textwidth]{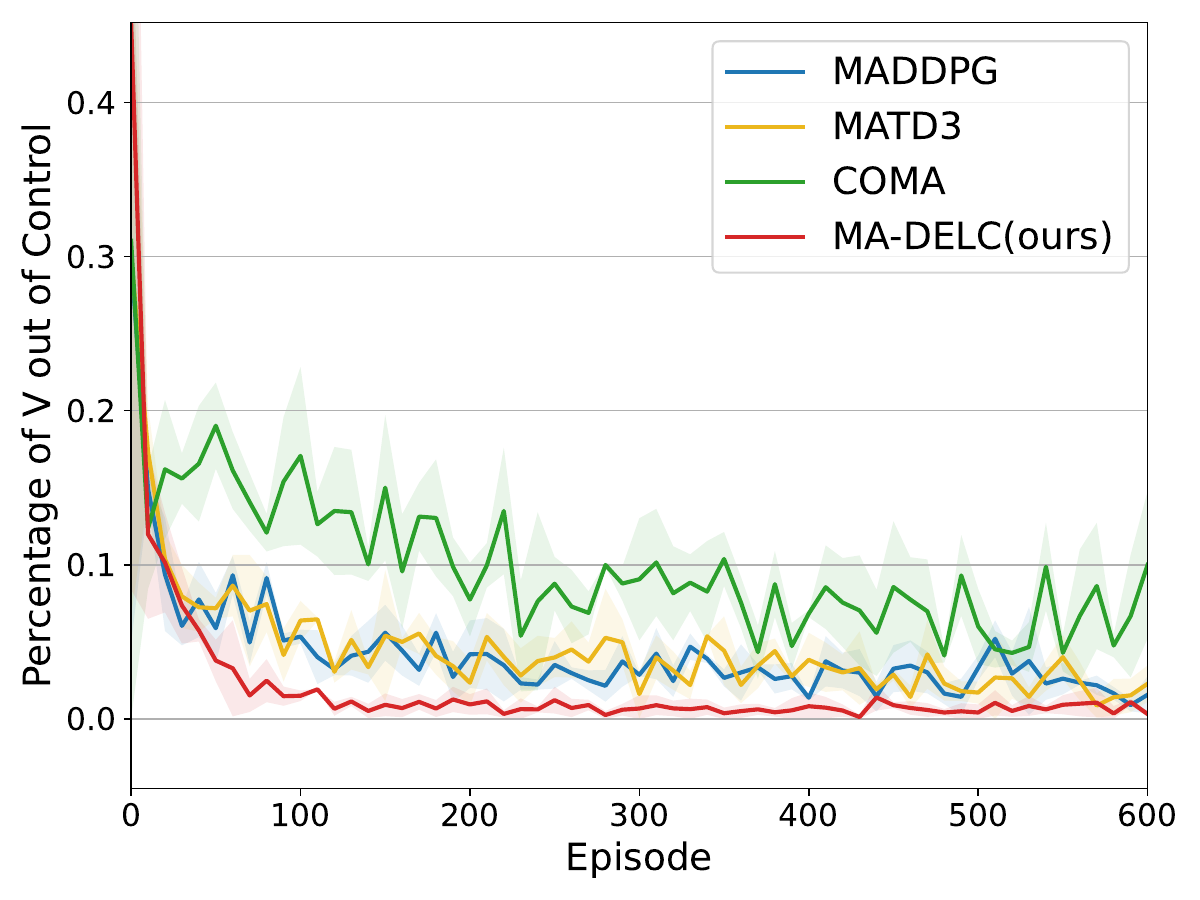}}
        \vspace{3pt}
 	\centerline{\includegraphics[width=\textwidth]{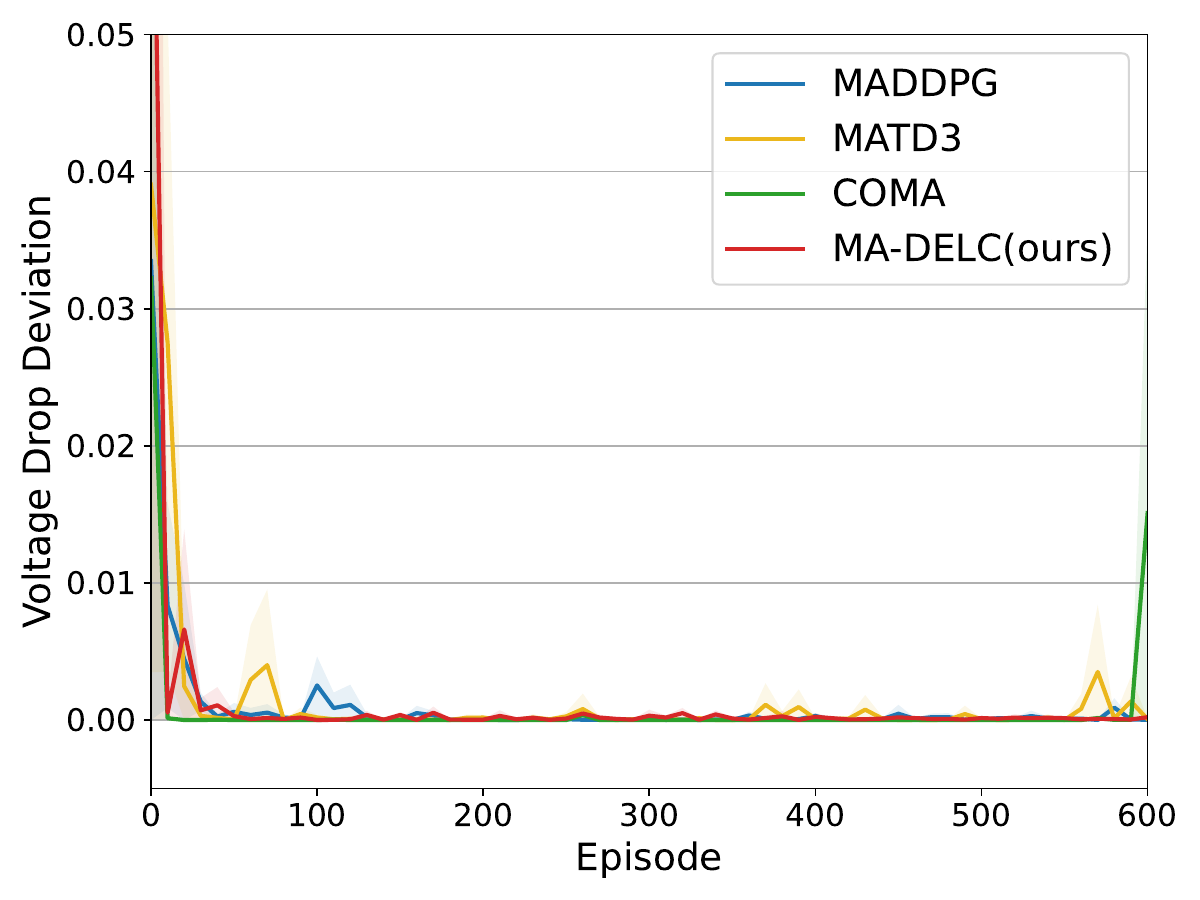}}
        \vspace{3pt}
 	\centerline{\includegraphics[width=\textwidth]{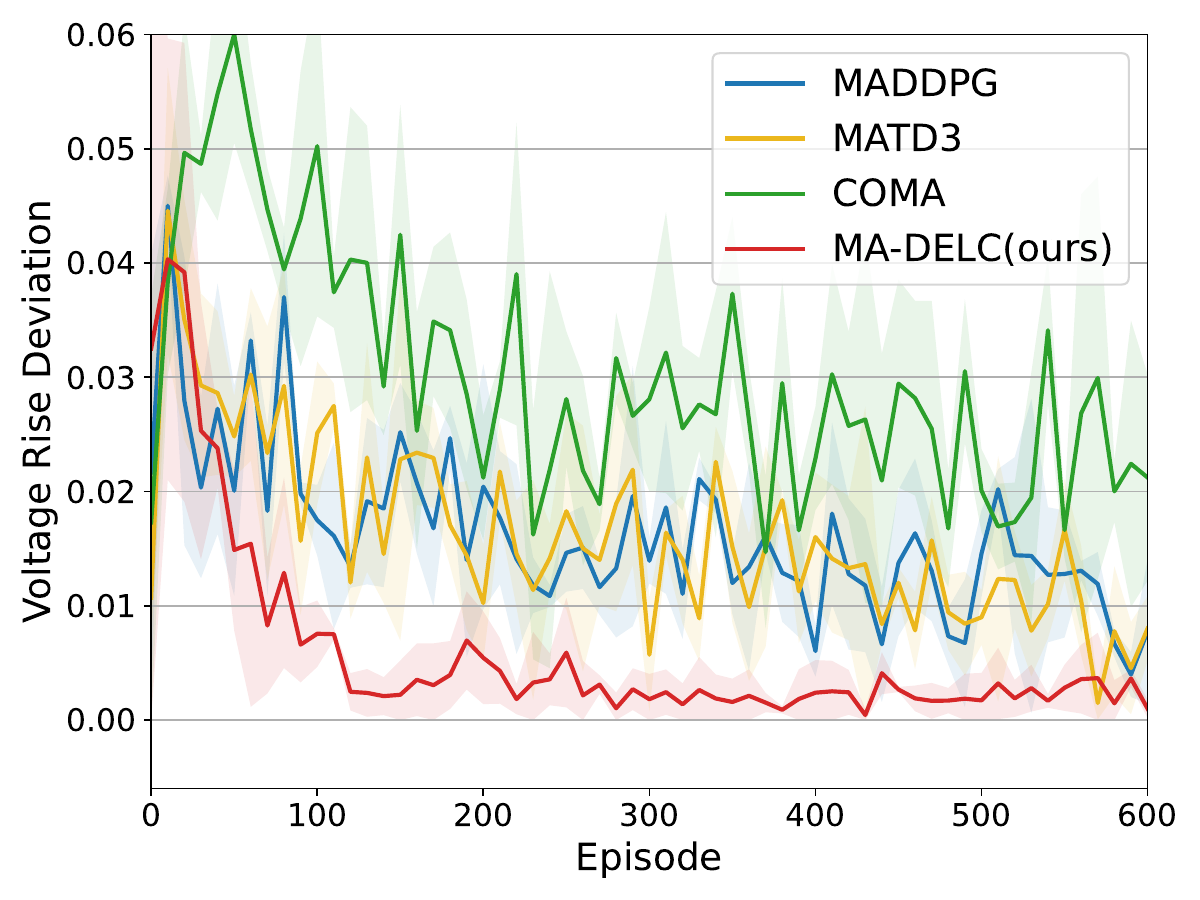}}
 	\vspace{3pt}
	\centerline{(c) 322-bus scenario}
 \end{minipage}
 \caption{Results on the fulfillment of the safety constraints of the different algorithms. Each row is the CR, Percentage of V out of Control, Voltage Drop Deviation, Voltage Rise Deviation metric respectively, and each column is the 33-bus, 141-bus, 322-bus scenario respectively. 
 }
 \label{fig:result_con}
\end{figure*}

\begin{figure*}[ht]	
 \begin{minipage}{0.32\linewidth}
 	\vspace{3pt}
 	\centerline{\includegraphics[width=\textwidth]{fig/QL-33bus}}
 	\vspace{3pt}
 	\centerline{\includegraphics[width=\textwidth]{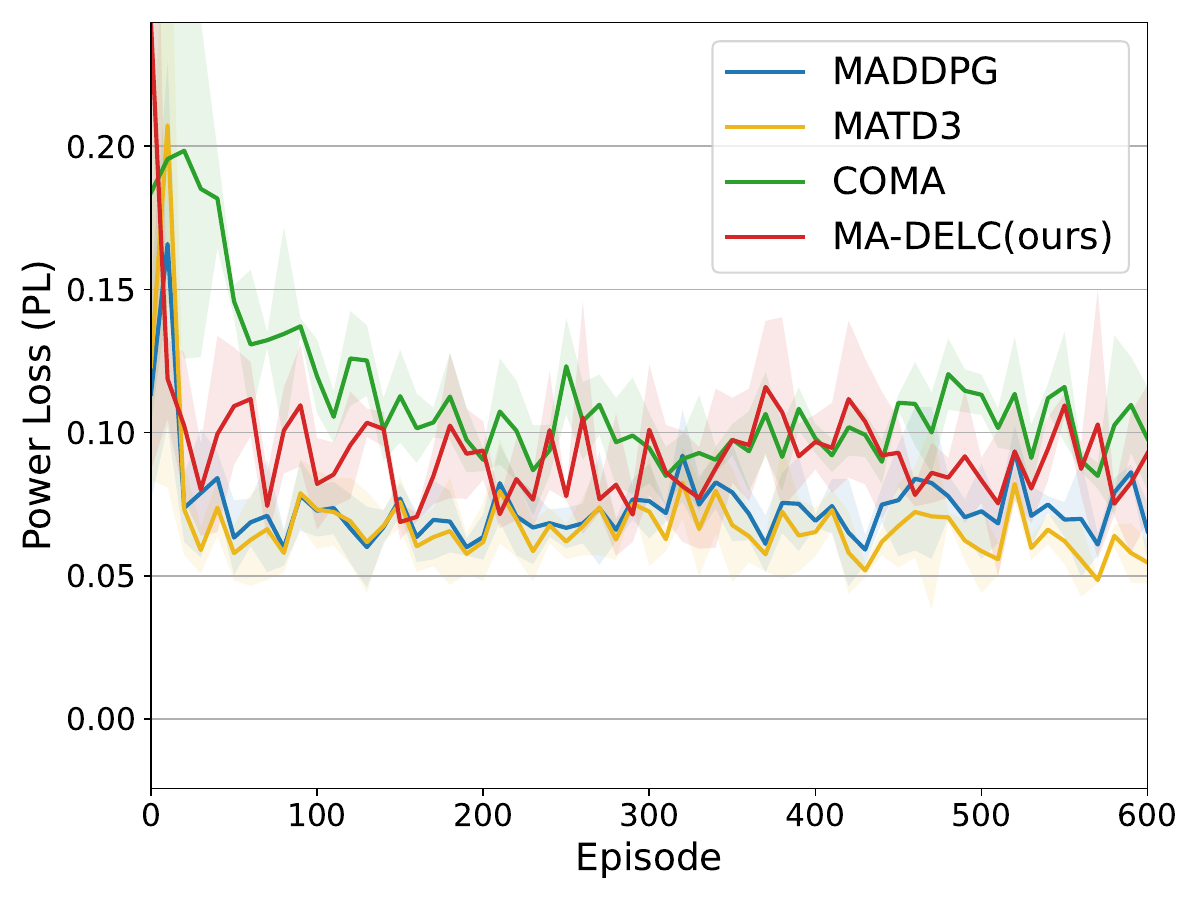}}
 	\vspace{3pt}
 	\centerline{(a) 33-bus scenario}
 \end{minipage}\hfill
 \begin{minipage}{0.32\linewidth}
	\vspace{3pt}
 	\centerline{\includegraphics[width=\textwidth]{fig/QL-141bus}}
 	\vspace{3pt}
 	\centerline{\includegraphics[width=\textwidth]{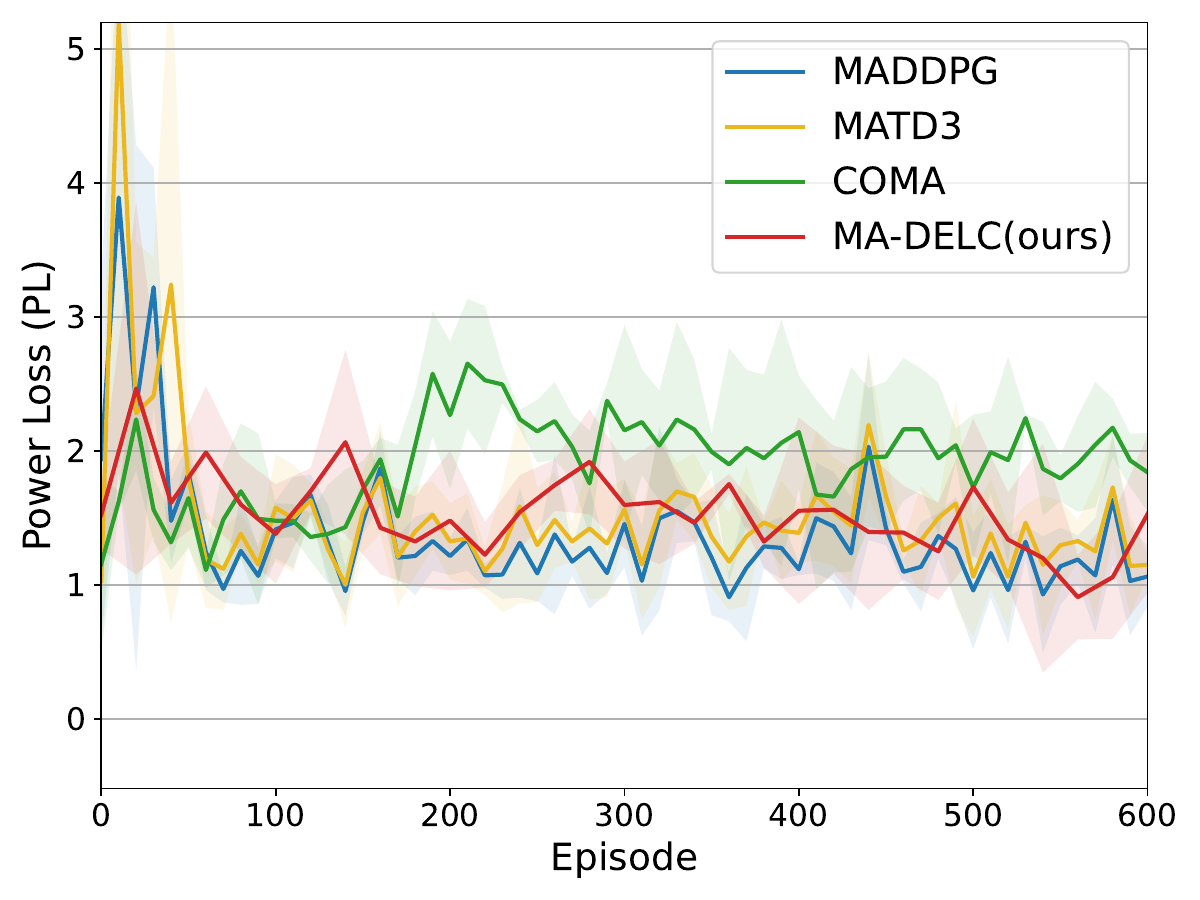}}
 	\vspace{3pt}
	\centerline{(b) 141-bus scenario}
\end{minipage}\hfill
\begin{minipage}{0.32\linewidth}
	\vspace{3pt}
 	\centerline{\includegraphics[width=\textwidth]{fig/QL-322bus}}
 	\vspace{3pt}
 	\centerline{\includegraphics[width=\textwidth]{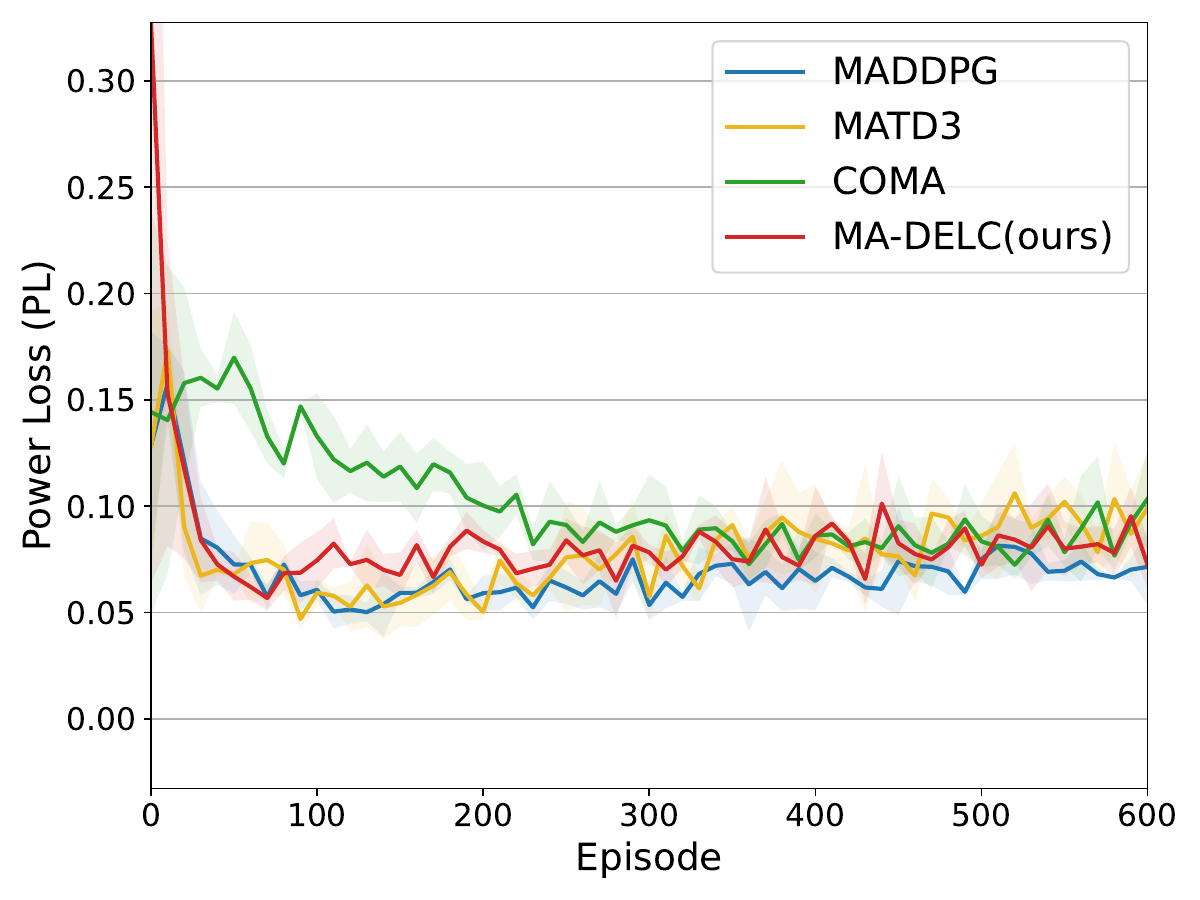}}
 	\vspace{3pt}
	\centerline{(c) 322-bus scenario}
 \end{minipage}
 \caption{Results on the power loss of the different algorithms. Each row is the QL and PL metric respectively, and each column is the 33-bus, 141-bus, 322-bus scenario respectively. 
 }
 \label{fig:result_obj}
\end{figure*}

During training, each episode starts at a random moment in a random day, and lasts for 12 hours (240 time steps). 
The algorithms update critic networks with 10 epochs while update actor networks with 1 epoch every 60 time steps. 
The target update learning rate is 0.01. The batch size of training data is 128 and the replay buffer size is 5000. 
We test the algorithm every 10 training episodes with 5 testing episodes. To avoid naive episodes in testing (e.g. an episode lasts from 6p.m. to 6a.m. the next day has little PV penetration and the agents are easy to control), 
each evaluation episode lasts for 24 hours (480 time steps). 
%We run each algorithm with 3 random seeds.

Based on the settings in \cite{nips2021MARLavc}, we adopt the reward below for non-constrained baseline methods:
\begin{equation}
        r=-\frac{1}{|V|}\sum_{i\in V}l_v(v_i)-\beta\cdot l_q(q^{PV})
\end{equation}
This reward function interwinds the optimization objective and the constraints. 
Here, $l_q(q^{PV})=\frac 1 {|I|} ||q^{PV}||$ is the reactive power generation loss of PVs --- an approximation of power loss related to agents. It is an approximation of the optimization objective of the active voltage control problem, and $l_q(q^{PV})$ alone is used as the reward for Lagrangian-constrained MARL method. 
$l_v(\cdot)$ is a voltage barrier function to measure voltage deviation from $1.0p.u.$. It is related to the voltage constraints. 
According to the experimental results of \cite{nips2021MARLavc}, the L1-shape voltage barrier function ($l(v)=|v-1.0|_1$) achieves the best CR among other voltage barrier functions. Hence we adopt this voltage barrier function in our experiments. 
$\beta$ is a hyper-parameter to trade-off between satisfying safety constraints and reducing power loss, and it serves as a fixed Lagrange multiplier. 
The manual choice of $\beta$ requires strong experience, and we adopt $\beta = 0.1$ following former settings.

Now, we introduce the hyperparameters of COMA \cite{COMA}, MADDPG \cite{MADDPG} and MATD3 \cite{MATD3} used in experiments. The variance $\Sigma_{std}=1.0$ of the Gaussian exploring action noise $\xi$ for MADDPG and our proposed method MA-DELC. The sample size M of COMA for continuous actions is set to 10 in experiments. The clip boundary $c$ for clipping the exploration noise for MATD3 is set to 1 in the experiments.

\section{Additional Experimental Results}
In this section, we evaluate our algorithm with additional metrics. Denote the voltage of node $i$ at step $t$ as $v_{i,t}$, and the max step of an evaluation episode as $T$, the metrics of accessing the fulfillment of the safety constraints are listed as follows. 
All of the following metrics are calculated for an episode.

\begin{itemize}
\item \textbf{Controllable Ratio (CR)}: It calculates the ratio of the steps in an episode where all buses' voltage is under control within the safety range. Note that higher CR is better and the optimal CR is 1.
\begin{equation}
\mbox{CR} =\frac{1}{T} \sum_{t=1}^T \I ( \!\!\! \sum_{i\in V\backslash \{0\}} \!\!\!\!\! \I(v_{i,t}<0.95 ~\mbox{or}~ v_{i,t}>1.05)=0 ).
\end{equation}
\item \textbf{Percentage of Voltage out of Control (PVooC)}: The percentage of V out of control in a certain step is the proportion of nodes whose voltages are out of the safe range in a step, and PVooC calculates the mean percentage of V out of control per step. CR focus on the proportion of controllable time steps, while PVooC focus on the proportion of controllable nodes. In other words, the PVooC metric indicates the spatial stability of voltage, whereas the CR metric indicates the temporal stability of voltage. We expect lower PVooC, and the optimal one is 0.
\begin{equation}
\mbox{PVooC} =\frac{1}{T} \sum_{t=1}^T \frac{1}{|V|}\sum_{i\in V\backslash \{0\}} \!\!\!\!\! \I(v_{i,t}<0.95 ~\mbox{or}~ v_{i,t}>1.05). 
\end{equation}
\item \textbf{Voltage Drop Deviation (VDD)}: It calculates the mean of the deviation of the lowest node voltage from $0.95p.u.$ per step. If all nodes' voltage exceed $0.95p.u.$, then the VDD of this step counts 0. Notice that large VDD may reduce the quality of power supply for users, even resulting in voltage collapse and large-scale power failure. We except lower VDD, and the optimal one is 0.
\begin{equation}
\mbox{VDD} =\frac{1}{T} \sum_{t=1}^T \max_{i\in V\backslash \{0\}} \{(0.95-v_{i,t}) \cdot \I(v_{i,t}<0.95)\}. 
\end{equation}
\item \textbf{Voltage Rise Deviation (VRD)}: Similarly, it calculates the mean of the deviation of the highest node voltage from $1.05p.u.$ per step. If all nodes' voltage are below $1.05p.u.$, then the VRD of this step counts 0. In real world, large VRD can negatively impact the performance and lifespan of electrical equipment, and even cause irreversible damage to the system equipment, increasing safety risks such as electric shock and fire hazards. We except lower VRD, and the optimal one is 0. 
\begin{equation}
\mbox{VRD} =\frac{1}{T} \sum_{t=1}^T \max_{i\in V\backslash \{0\}} \{(v_{i,t}-1.05) \cdot \I(v_{i,t}>1.05) \}. 
\end{equation}
Note that both VDD and VRD give a quantitative description of constraint violation, which are critical for the voltage control domain.
\end{itemize}

These metrics allow for a comprehensive assessment of the algorithms' ability of fulfilling the constraints. Results about the fulfillment of the constraints are shown in Figure~\ref{fig:result_con}. 

In both the 33-bus and 141-bus scenario, the CR of MA-DELC is close to 1. As expected, the PVooC, VDD and VRD of our MA-DELC method in these two scenarios are close to 0. 
In the larger 322-bus scenario, the CR of MA-DELC is about 0.9. As we can see, the PVooC of MA-DELC is the lowest comparing to the other baselines. Meanwhile, the variance of PVooC of MA-DELC is relatively small. These additional results emphasize the stability of our method, despite that it is difficult to reach the CR of 1 in the 322-bus scenario. 
Note that CR and PVooC have different emphasis and the relationship between the two is not simply added to 1. 
In the 322-bus scenario, the CR of COMA after convergence is similar to MADDPG. However, PVooC of COMA is higher than MADDPG, meaning that COMA is less stable than MADDPG in this scenario. 

One of the main difficulties of the active voltage control problem is to keep the voltage stable with the high penetration of PVs. Therefore, the VRD is generally more difficult to regulate than the VDD. 
As shown in the experiments, this is more obvious in the 322-bus scenario, where the VDD of each methods is close to 0.0 but the VRD varies. 
Remarkably, our MA-DELC approach outperforms the other baselines regarding the VRD metric in the 322-bus scenario. 
In summary, results of Figure~\ref{fig:result_con} confirm the effectiveness of our method in regulating voltage --- the primary concern of this domain.

The metrics of accessing the optimization of objective (i.e. reducing the total power loss) are listed as below. Note that both of QL and PL are the lower the better.
\begin{itemize}
\item \textbf{Q Loss (QL)}: It calculates the mean reactive power generations by agents per time step. This metric is chosen as a proxy for power loss (PL), which is often challenging to obtain for the entire network. QL serves as an approximation of the power loss attributable to PVs (agents).
\begin{equation}
    \mbox{QL} =\frac{1}{T}\sum_t \frac 1 {|\mathcal{N}|} \sum_{i\in\mathcal{N}} \left\| q^{PV}_{i, t} \right\|
\end{equation}
\item \textbf{Power Loss (PL)}: It calculate the mean of the total line loss of the network per time step. It is the exact power loss (i.e. the optimization object of the constrained optimization problem of active voltage control) as in Equation 1. However, it is unavailable without the physical model of the power distribution network, and is often challenging to obtain in real time. 
\begin{equation}
    \mbox{PL} = \frac{1}{T} \sum_t \sum _{(i,j)\in E}r_{ij}l_{ij,t}
\end{equation}
where $r_{ij}$ is the resistance on branch $(i,j)$, which belongs to the specific physical parameters of the power network model, and $l_{ij,t}$ is the square of the current magnitude from bus $i$ to $j$ at time step $t$. 
In our  experiments, $l_{ij,t}$ is determined by the simulator solving the physical constraints outlined in Equation~\ref{eq:complete_avc}. This is unavilable in the model-free RL settings. 
In contrast, QL can be readily calculated with the penetration power and control of PVs. Therefore, QL is used as the reward of RL.
\end{itemize}

Our results are shown in Figure~\ref{fig:result_obj}. 
As we can see, the results of QL and PL is not always consistent. For example, in the 450-600 episode of the 33-bus scenario, the QL of MA-DELC is higher than COMA but the PL of MA-DELC is lower than COMA. 
This is because that QL simply calculates the mean loss of each agent, but PL needs to take the precise power network model into consideration. 
In the 33-bus scenario, our method MA-DELC has slightly higher QL than the other baselines. Specifically, it has moderate PL among the other methods. In the 141-bus scenario, MA-DELC achieves QL and PL similar to the lowest values of the baselines. In the 322-bus scenario, the QL and PL of all the methods are rather small and close to each other.
In conclusion, the experimental results of these additional metrics in Figures \ref{fig:result_con} and \ref{fig:result_obj} further verify the effectiveness of our proposed method in fulfilling the constraints and optimizing the objective concurrently.

\end{document}